\documentclass[article,onecolumn,11pt]{IEEEtran}
\IEEEoverridecommandlockouts
\usepackage{multirow}
\usepackage{cite}
\usepackage{amsmath,amssymb,amsfonts}
\usepackage{algorithmic}
\usepackage{graphicx}
\usepackage{textcomp}
\usepackage{xcolor}
\usepackage{fullpage}
\usepackage{framed}

\newcommand{\Val}{\textnormal{Val}}
\newcommand{\Set}{\textnormal{Set}}
\newcommand{\Var}{\textnormal{Var}}

\newcommand{\Fus}{\mathbf{F}}
\newcommand{\PrL}{{\Pr}_L}
\newcommand{\PrU}{{\Pr}_U}
\newcommand{\Bel}{\textnormal{Bel}}
\newcommand{\Pl}{\textnormal{Pl}}
\newtheorem{defn}{Definition}

\newtheorem{prob}{Problem}
\def\BibTeX{{\rm B\kern-.05em{\sc i\kern-.025em b}\kern-.08em     T\kern-.1667em\lower.7ex\hbox{E}\kern-.125emX}}

\begin{document}

\title{Information Fusion  on Belief Networks}

\author{\IEEEauthorblockN{Shawn C. Eastwood}\\
\IEEEauthorblockA{\textit{Wentworth Institute of Technology, Boston, USA}}\\
\and
\IEEEauthorblockN{Svetlana N. Yanushkevich}\\
\IEEEauthorblockA{\textit{Department of Electrical \& Computer  Engineering, University of Calgary, Canada}
}
}

\maketitle

\begin{abstract}
This paper will focus on the process of ``fusing" several observations or models of uncertainty into a single resultant model. Many existing approaches to fusion use subjective quantities such as ``strengths of belief" and process these quantities with heuristic algorithms. This paper argues in favor of quantities that can be objectively measured, as opposed to the subjective ``strength of belief" values. This paper will focus on probability distributions, and more importantly, structures that denote sets of probability distributions known as ``credal sets". The novel aspect of this paper will be a taxonomy of models of fusion that use specific types of credal sets, namely probability interval distributions and Dempster-Shafer models. An objective requirement for information fusion algorithms is provided, and is satisfied by all models of fusion presented in this paper. Dempster's rule of combination is shown to not satisfy this requirement. This paper will also assess the computational challenges involved for the proposed fusion approaches.
\end{abstract}

%\begin{keyword}
%information fusion \sep uncertain probabilities \sep credal sets \sep probability interval distributions \sep Dempster-Shafer theory \sep computational challenges 
%\end{keyword}

\section{Introduction}

The problem of ``fusion" stems from the need to combine information from various sources. Each of these source is assumed to provide either an observation, or a ``model of uncertainty". Several approaches to fusing uncertainty models have been investigated in \cite{[Delmotte2004]} (the transferable belief model),
%\cite{[Yager2004]},
\cite{[Yager2004II]} (uninorm aggregation), %\cite{[Yager2011]}, %and \cite{[Yager2016IF]}. 
\cite{[Yager1987]} (Dempster-Shafer fusion), and
\cite{[Dezert2003],[Norden2008],[Tchamova2003]} (Dezert-Smarandache fusion).  
A survey of contemporary fusion approaches is given in \cite{[Khaleghi2013]}. Most of these approaches however rely on quantities such as ``strengths of belief" that are highly subjective. In addition, the heuristics used to handle strengths of belief are algorithms designed so that the outputs ``make sense" as opposed to obeying an objective criteria. 

This paper will utilize convex sets of probability distributions, known as credal sets, to form the model of uncertainty in probability distributions. However, unlike most approaches to credal sets which maintain a list of the extreme points, this paper will focus on ``subtypes" of credal sets, namely probability interval distributions and Dempster-Shafer models (Dempster-Shafer models can also describe credal sets in addition to heuristic belief functions, as will be discussed in section \ref{sec:DS_Fusion}). The fusion of credal sets as models of uncertainty is described in \cite{[Karlsson2010],[Karlsson2011],[Karlsson2013]}, and it is already established that fusion can be performed in an efficient and exact manner when credal sets are denoted by listing their extreme points. It is known however, that credal set subtypes such as probability interval distributions and Dempster-Shafer models can describe certain credal sets using dramatically less data than the number of extreme points (see \cite{[Tessem1992], [Zaffalon2002]} for a discussion on the number of extreme points of probability interval distributions). When a credal set is the set of probability distributions denoted by a specific probability interval distribution or Dempster-Shafer model, the probability interval distribution or Dempster-Shafer model is the more efficient representation in terms of space (and subsequently computational complexity). This is the prime motivation for using the credal set subtypes of probability interval distributions and Dempster-Shafer models, and gives practical purpose to the catalog of fusion approaches presented in this paper.

The use of intervals to describe probabilities is formally developed in \cite{[DeCampos1994]} and \cite[chapter 5]{[Klir2005]} and is described in detail in section \ref{sec:probability_intervals}.

Dempster-Shafer (DS) theory is described in \cite[chapter 5]{[Klir2005]}, \cite{[Yager1987]} and section \ref{sec:DS_Fusion}. In many publications such as \cite{[Yager1995IEEE]}, Dempster-Shafer models are interpreted as follows: the belief and plausibility respectively form lower and upper bounds for the true probability. This is the interpretation of Dempster-Shafer theory that will be used in this paper. With this interpretation, Dempster-Shafer models effectively denote a credal set. Dempster's rule of combination, described in \cite[chapter 5]{[Klir2005]} and \cite{[Yager1987]}, is a popular approach to fusing Dempster-Shafer models. There is however, a major inconsistency with Dempster's rule of combination when Dempster-Shafer models are interpreted as credal sets: the fused Dempster-Shafer model does not describe a credal set that contains all possible probability distributions that result from fusing probability distributions chosen from the credal set of each input Dempster-Shafer model (see definition \ref{def:containment_property_general_fusion}). This inconsistency is described in section \ref{sec:conjunctive_combination}.

Many approaches to fusion using Dempster-Shafer models focus on ``redistributing conflict" such as from \cite{[Dezert2006],[Smarandache2005IF],[Smarandache2005]}. ``Conflict redistribution" focuses on minimizing or eliminating the renormalization that occurs when Dempster-Shafer models are combined/fused. This paper, due to its focus on credal sets, will not focus on approaches such as conflict redistribution, since these approaches do not treat Dempster-Shafer models as credal sets.  

%The use of intervals to describe probabilities is formally developed in \cite{[DeCampos1994]} and \cite[chapter 5]{[Klir2005]}. In publications such as \cite{[Cano2007], [Cozman2005], [Zaffalon2002]}, convex sets of probability distributions, referred to as ``credal sets", are utilized to quantify uncertainty in probability distributions. \cite{[Cano2007], [Cozman2005],[Karlsson2010],[Karlsson2011],[Karlsson2013], [Zaffalon2002]} give discussions of probabilistic inference using credal sets, and this paper will contribute to the discussion by examining specific ``types" of credal sets, and applying such interpretations to two different approaches to information fusion. Two specific types of credal sets will be examined: probability interval distributions and Dempster-Shafer models. The computational challenges that arise from the use of credal sets will be addressed to yield fusion rules that satisfy objective criteria, as well as computational tractability.

%The key contributions of this paper are as follows: the identification of two specific modes of information fusion; the close examination of the credal set sub-types of probability intervals and Dempster-Shafer models; a taxonomy of fusion approaches and algorithms that have varying levels of computational complexity and accuracy; and an emphasis on the need for objective measures and criteria for information fusion.

An important aspect of this paper is a look at the various algorithms that fuse credal sets. Existing work on this topic include the known Bayesian fusion of credal sets from \cite{[Karlsson2011]}, the calculation of posterior probability intervals from \cite{[DeCampos1994], [Walley1996], [Antonucci2013ECSQARU]}, and the creation of software packages that calculate posterior credal sets such as ``CREDO" \cite{[Antonucci2013]}. This paper will propose a catalog of fusion approaches that utilize probability interval distributions and Dempster-Shafer models. This taxonomy will include existing work and algorithms generated specifically for this paper.  

The structure of this paper is as follows: Section \ref{sec:Background} will review two different modes of fusion using point probability distributions. Section \ref{sec:Complex} will review the requirements for fusion involving sets of probability distributions. Section \ref{sec:probability_intervals} will cover the use of probability intervals. Section \ref{sec:DS_Fusion} will cover the use of Dempster-Shafer models, and propose an alternative to Dempster's rule of combination.

\section{Contributions}\label{sec:Contributions}

The contributions of this paper are: 
\begin{itemize}
\item The most important contribution of this paper is a taxonomy and catalog of fusion approaches and algorithms that utilize ``subtypes" of credal sets, in this case ``probability interval distributions" and ``Dempster-Shafer models". All fusion approaches will satisfy an important objective criteria, referred to in this paper as the ``containment property". Special attention is paid to the computational challenges involved. Various approaches are given, which exhibit trade-offs between accuracy and computational complexity. %In addition, an NP-hard problem is identified and shown to be NP-hard. 
Some of the fusion approaches are already known to the literature (such as context specific fusion with probability intervals described in \cite{[Walley1996]}), and others were created specifically for this paper. 
%\item Credal sets are a known highly expressive approach to denoting sets of probability distributions. The utility of probability interval distributions and Dempster-Shafer models as special subtypes of credal sets is demonstrated using the criteria of space (memory) complexity (see sections \ref{sec:prob_intervals_credal_sets} and \ref{sec:DS_credal_sets}). This in turn motivates use of the algorithms cataloged by this paper.
\item A proposed objective criteria for information fusion referred to as the ``containment property" (see section \ref{sec:Complex} for the definitions) is given. Dempster's rule of combination is shown to violate the containment property.
\item A distinction is made between two types of information fusion, referred to as ``context specific" and ``general fusion". Each type of fusion has different information requirements, and the algorithms are different. Context specific fusion requires more prior information, but is less computationally intensive than general fusion. The important distinction between context specific fusion and general fusion is that context specific fusion only requires raw observations as input, while general fusion requires complete credal sets. Context specific fusion follows the hypothesis-observation models used in publications such as \cite{[Delmotte2004],[Zaffalon2002]} and \cite[section 4, calculus]{[Walley1996]}, and the algorithms are generally polynomial time with respect to the size of the input. General fusion is similar to the direct Bayesian fusion of credal sets. While the direct Bayesian fusion of credal sets can be performed exactly in polynomial time \cite[Theorem 2]{[Karlsson2011]}, when credal sets are restricted to specific subtypes, general fusion becomes much more difficult.   
\end{itemize}

\section{Background}\label{sec:Background}

%************************ Uncertainty Structures ***************************

In the literature, 
%\cite{[Cano2007], [Cozman2000], [Cozman2005], [Karlsson2010], [Karlsson2011], [Karlsson2013], [Zaffalon2002]}, 
convex sets of probability distributions are referred to as ``credal sets". 
%This paper, however, will use the more intuitive terminology of ``uncertainty structure".

\begin{defn}
A \textbf{credal set}, is a convex set of probability distributions. All probability distributions in a credal set cover the same variables and have the same domain.

Specific ``subtypes" of credal sets that will be the focus of investigation include ``probability interval distributions" and ``Dempster-Shafer models". In \cite{[Karlsson2011]}, credal sets are denoted by listing their ``extreme points". The extreme points are points that belong to the credal set, but are not a convex combination of other points in the credal set \cite{[Karlsson2011]}. Each subtype of credal set however has a more compact style of representation that comes with the restriction that there are some credal sets that cannot be represented by the current subtype. 

In this paper, a credal set subtype is considered to be ``non-trivial" if and only if it denotes a set of probability distributions as opposed to a single probability distribution. 
\end{defn}

The following notation will be used with respect to credal sets:
\begin{itemize}
\item Given a single variable \(x\), the set \(\{x\}\) will be denoted by simply using \(x\).
\item Given a set of variables \(X\), \(\Val(X)\) is the set of all possible complete assignments to the variables in \(X\).
\item Given a set of variables \(X\), the set of sets \(2^{\Val(X)} \setminus \{\emptyset\} = \{A \subseteq \Val(X) : A \neq \emptyset\}\) is denoted by \(\Set(X)\).
%\item Given a single probability distribution \(\Pr\), the set \(\{\Pr\}\) will be denoted using simply \(\Pr\).
\item Given a credal set \(S\), 
	\begin{itemize}
	%\item The set of all probability distributions contained by \(S\) is denoted by simply \(S\).
	\item \(\Var(S)\) is the set of variables covered by each probability distribution from \(S\).
	\item \(\Val(S)\) denotes \(\Val(\Var(S))\).
	\item \(\Set(S)\) denotes \(\Set(\Var(S))\).
	\end{itemize}
\item Given an arbitrary condition \(C\), 
	\begin{itemize}
	\item \(\Pr_L(C) = \min(\Pr(C))\) denotes the smallest probability that \(C\) is satisfied.   
	\item \(\Pr_U(C) = \max(\Pr(C))\) denotes the largest probability that \(C\) is satisfied.   
	\end{itemize}
\end{itemize}

In addition, pseudo code will be used to describe various algorithms. Comments in the pseudo code are denoted using a double forward slash: \texttt{//}, or are enclosed by: \texttt{/* ... */}.

%********************* Fusion Problem ************************************

\subsection{Two Approaches to Fusion}

In this paper, fusion will occur within the context of trying to identify an object using information gathered by remote sensors.

There are two fusion problems that will be considered by this paper:
\begin{prob}
\textbf{Context Specific Fusion}: Consider a variable of interest, hypothesis variable \(H\). Given several (\(N \geq 1\)) observations \(O_1, O_2, \dots, O_N\), we wish to generate a ``posterior" credal set \(S_\bullet = \Fus(O_1, O_2, \dots, O_N)\) that covers the hypothesis variable \(H\). \(S_\bullet\) should consolidate all of the observations \(O_1, O_2, \dots, O_N\). The hypothesis variable, \(H\), will be assumed to have \(M \geq 2\) possible values, denoted by: \(1, 2, \dots, M\). The subtype of the posterior credal set will be the same as the subtype of the credal set that is the ``prior" for \(H\). 

\textbf{General Fusion}: Consider a variable of interest, hypothesis variable \(H\). Given several (\(N \geq 2\)) credal sets \(S_1, S_2, \dots, S_N\) that cover the hypothesis variable \(H\), we wish to generate a ``posterior" credal set \(S_\bullet = \Fus(S_1, S_2, \dots, S_N)\) that covers \(H\) and consolidates all of the information from \(S_1, S_2, \dots, S_N\). The hypothesis variable, \(H\), will be assumed to have \(M \geq 2\) possible values, denoted by: \(1, 2, \dots, M\). The subtype of the posterior credal set will be the same as the subtype of the input credal sets. 
\end{prob}

\begin{figure}[h!]
\begin{centering}
\begin{tabular}{cc}
\includegraphics[width = 0.5\textwidth]{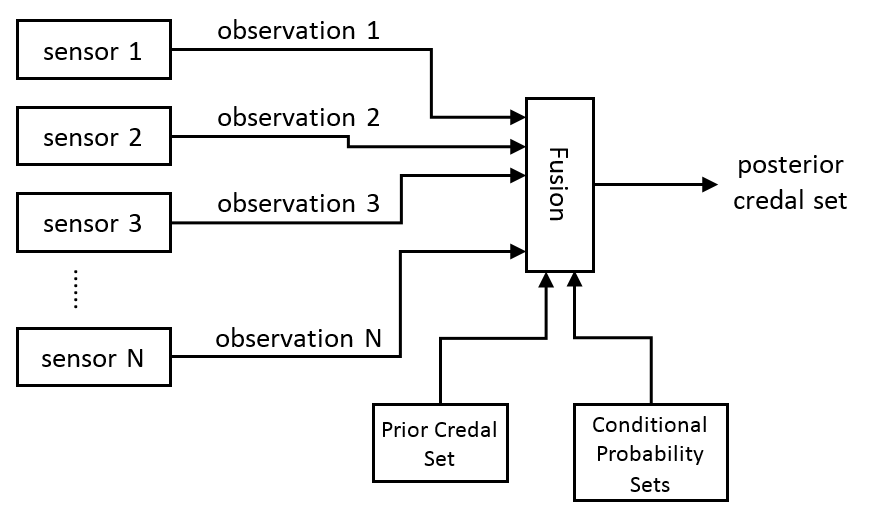} & \includegraphics[width = 0.5\textwidth]{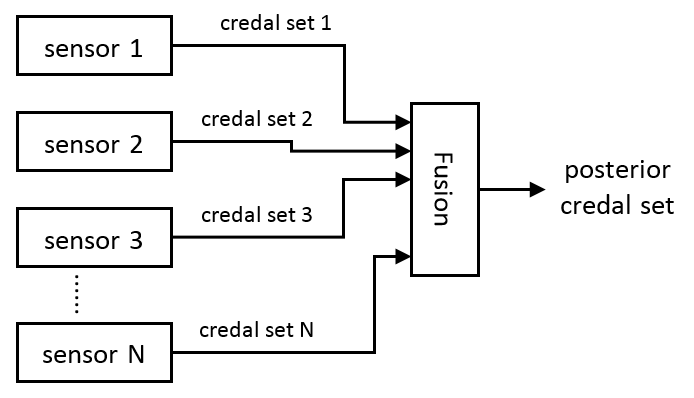} \\
(a) & (b) \\
\end{tabular}
\caption{(a) The process of Context Specific Fusion. (b) The process of General Fusion.}
\label{fig:Fusion_Types}
\end{centering}
\end{figure}

The process of context specific fusion is shown in figure \ref{fig:Fusion_Types}(a). Observations \(O_1, O_2, \dots, O_N\) are acquired from various sources: in this case, the sources are sensors. Alongside existing data in the form of a prior credal set for \(H\) and probability ranges for each observation given each possible value of \(H\), the observations are fused to produce a ``posterior" credal set that describes \(H\). This is the approach to fusion used in \cite{[Delmotte2004],[Zaffalon2002]} and \cite[section 4, calculus]{[Walley1996]}.

The process of general fusion is shown in figure \ref{fig:Fusion_Types}(b). ``Prior" credal sets \(S_1, S_2, \dots, S_N\) are acquired from various sources: in this case, the sources are sensors. The credal sets are fused to produce a ``posterior" credal set that describes \(H\). Unlike context specific fusion, general fusion does not require existing data. Despite this however, each sensor must be equipped with the capacity to return a credal set about \(H\), as opposed to raw data and observations. This is the approach to fusion used in many publications such as \cite{[Guo2010], [Karlsson2011], [Yager2004II], %[Yager2011], 
[Yager2016IF]}.

To describe in detail how each fusion process works, the concept of causal networks needs to be established:

%***************** Causal Network ******************************

\begin{defn}
A \textbf{Causal Network} is a directed acyclic graph wherein random variables are depicted as nodes. A directed edge from node \(x\) to \(y\) indicates that variable \(x\) has a direct/causal influence on \(y\). Without any conditioning of any random variables, the model of uncertainty (probability distribution, credal set, or other model) that describes \(y\) is a function of the set of ``parents" of \(y\). A parent of \(y\) is a variable like \(x\) which is the source of an edge that terminates on \(y\). Unlike Bayesian networks, the precise nature of a child node's dependency of the values of the parent node does not need to be specified in a causal network. 
\end{defn}

When conditional probability tables are used to describe how each node depends on its parents, the causal network becomes the well-known ``Bayesian network".

When the probability distributions in the conditional probability tables in a Bayesian network are replaced with credal sets, the result is a ``credal network". Theory related to credal networks can be found in \cite{[Cozman2000]}. An important concept related to credal networks is the concept of the ``strong extension", which is the tightest convex hull that contains all joint probability distributions allowed by the credal network.

The fact that in ``causal networks", the manner of a child node's dependency on its parent does not need to be specified, means that causal networks can include Bayesian networks and credal networks.

A causal network provides a concrete high level model of the scenario of interest. In the case of fusion, there will be a distinct causal network for each fusion approach.

\begin{figure}[h!]
\begin{centering}
\begin{tabular}{cc}
\includegraphics[width = 0.5\textwidth]{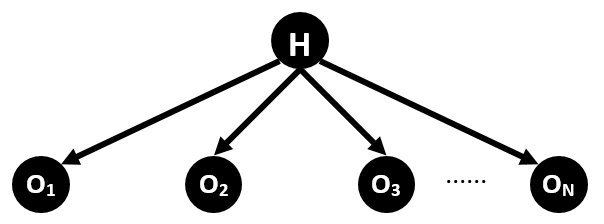} & \includegraphics[width = 0.5\textwidth]{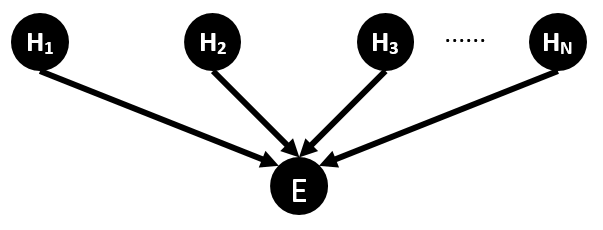} \\
(a) & (b) \\
\end{tabular}
\caption{(a) The causal network that describes the scenario envisioned for context specific fusion. (b) The causal network that describes the scenario envisioned for general fusion.}
\label{fig:Fusion_Causal_Network}
\end{centering}
\end{figure}

Figure \ref{fig:Fusion_Causal_Network}(a) displays the causal network that describes the scenario used for context specific fusion. In this scenario, the hypothesis variable \(H\) influences each of the observations \(O_1, O_2, \dots, O_N\). For each possible hypothesis, the observations all occur independently (there are no causal links between observations). 

Figure \ref{fig:Fusion_Causal_Network}(b) displays the causal network that describes the scenario used for general fusion. Unlike the causal network for context specific fusion, there are instead \(N\) hypothesis variables \(H_1, H_2, \dots, H_N\) that correspond to each of the \(N\) credal sets \(S_1, S_2, \dots, S_N\). The hypothesis variables are all independent (there are no causal links between the \(H_i\)'s). There is then a binary variable \(E\) which attains 1 if and only if all of the hypothesis variables are equal and 0 if otherwise.

%The causal network that relates variables \(H_1, H_2, \dots, H_N\) and \(E\) is shown in figure \ref{fig:Fusion_Causal_Network_2}. Each \(H_i\) is described by \(S_i\), and \(E = 1\) if \(H_1 = H_2 = \dots = H_N\); and \(E = 0\) if otherwise.

%********************** Conditioning ****************************

%The process of conditioning variables or random variables is crucial to the process of deriving conclusions from causal networks, for both context specific and general fusion. During conditioning, some variables are fixed to a predetermined value. These restrictions impact other variables. To demonstrate the effect of conditioning, consider binary variables \(A, B\), and \(C\). Let \(\Pr(A = 1) = 0.7\); \(\Pr(B = 1) = 0.5\); and \(C = A \vee B\), where \(\vee\) is the logical ``OR" operation. Imagine that we now force the condition \(C = 1\). The only possible states for \(A\), \(B\) and \(C\) together are \((A,B,C) =\) (0,1,1); (1,0,1); and (1,1,1). The ``prior" probabilities of these 3 states were originally 0.15; 0.35; and 0.35 respectively. Since all probabilities must sum to 1 however, these probabilities must be renormalized to give the ``posterior" probabilities 3/17; 7/17; and 7/17 respectively. As can be seen, the probability distribution over \(A\) and \(B\) is impacted by the information that \(C = 1\).

In this paper, the term ``prior" refers to probabilities before the values of any of the variables are known; and the term ``posterior" refers to probabilities after the values of certain variables have become known.

During context specific fusion, the observation variables \(O_1, O_2, \dots, O_N\) are fixed to their observed values, and the posterior credal set for \(H\) is computed. During general fusion, the binary variable \(E\) is fixed to 1 which forces all of the hypothesis variables to have the same value. The posterior credal set that describes this common value is the resultant posterior credal set for \(H\).

%************************** Point probabilities **********************

\subsection{Context Specific Fusion using point probabilities}\label{sec:probability_cs}

This section describes the process of context specific fusion when the credal sets consist of single probability distributions. A credal set that contains a single probability distribution is referred to as a ``point probability distribution". Before any fusion can occur, a ``prior" probability distribution for \(H\) is required. Let the prior probability of \(H = j\) for each \(j = 1, 2, \dots, M\) be denoted by \(p_j\). In addition, for each \(O_i\) (\(i = 1, 2, \dots, N\)), an observation \(o_i\) is received. For each \(j = 1, 2, \dots, M\), \(p_{i,j}\) will denote the probability of \(O_i = o_i\) provided that \(H = j\): \(p_{i,j} = \Pr(O_i = o_i|H = j)\). 
%When the condition \(O_i = o_i\) is enforced for each \(i = 1, 2, \dots, N\), the only possible states that the entire ensemble of variables \((H, O_1, O_2, \dots, O_N)\) can attain are:
%\[(H, O_1, O_2, \dots, O_N) = (1, o_1, o_2, \dots, o_N); (2, o_1, o_2, \dots, o_N); \dots; (M, o_1, o_2, \dots, o_N)\]
%The prior probabilities of these states are respectively:
%\[\forall j \in \{1, 2, \dots, M\} : \Pr(j, o_1, \dots, o_N) = p_j\prod_{i=1}^N p_{i,j}\]
%After renormalization, the posterior probabilities are respectively:
%\[\forall j' \in \{1, 2, \dots, M\} : \Pr(j', o_1, \dots, o_N) = \frac{p_{j'}\prod_{i=1}^N p_{i,j'}}{\sum_{j=1}^M p_j\prod_{i=1}^N p_{i,j}}\]
Bayes' rule gives gives the following posterior probability distribution for \(H\):
\[\forall j' \in \{1, 2, \dots, M\}: \Pr(H = j'|\forall i \in \{1, 2, \dots, N\} : O_i = o_i) = \frac{p_{j'}\prod_{i=1}^N p_{i,j'}}{\sum_{j=1}^M p_j\prod_{i=1}^N p_{i,j}}\]

\begin{framed} % *************** begin example ******************
As an example of context specific fusion, consider a machine that can be in one of 2 states: ``functional"; or ``non-functional". The machine's state is the hypothesis variable \(H\), and the \(M = 2\) states are respectively enumerated by 1  and 2. Imagine there are \(N = 3\) sensors in place to determine the state of the machine: sensor 1 (\(O_1\)) can return either ``low temperature" or ``high temperature"; sensor 2 (\(O_2\)) can return either ``low load" or ``high load"; and sensor 3 (\(O_3\)) can return either ``low current" or ``high current". Through careful experimentation, it is known that:
\begin{align*}
\Pr(H = 1) &= 0.9                                                          & \Pr(H = 2) &= 0.1 \\
\Pr(O_1 = \text{``low temperature"}|H = 1) &= 0.9  & \Pr(O_1 = \text{``low temperature"}|H = 2) &= 0.4 \\
\Pr(O_2 = \text{``low load"}|H = 1) &= 0.3               & \Pr(O_2 = \text{``low load"}|H = 2) &= 0.6 \\
\Pr(O_3 = \text{``low current"}|H = 1) &= 0.7          & \Pr(O_3 = \text{``low current"}|H = 2) &= 0.2 
\end{align*}
% a functional machine exhibits a ``low temperature" 90\% of the time, a ``low load" 30\% of the time, and a ``low current" 70\% of the time; 
%and that a non-functional machine exhibits a ``low temperature" 40\% of the time, a ``low load" 60\% of the time, and a ``low current" 20\% of the time. 
%It is also known that without the information of any sensors, that the machine is functional 90\% of the time.

If sensor 1 returns \(O_1 = \)``high temperature"; sensor 2 returns \(O_2 = \)``low load"; and sensor 3 returns \(O_3 = \)``low current"; then the posterior probability distribution for \(H\) is:
\[\Pr(H = 1|O_1, O_2, O_3) = \frac{(0.9)(0.1)(0.3)(0.7)}{(0.9)(0.1)(0.3)(0.7) + (0.1)(0.6)(0.6)(0.2)} \approx 0.7241\]
\[\Pr(H = 2|O_1, O_2, O_3) = \frac{(0.1)(0.6)(0.6)(0.2)}{(0.9)(0.1)(0.3)(0.7) + (0.1)(0.6)(0.6)(0.2)} \approx 0.2759\]
\end{framed} % ******************** end example *****************

It should also be noted that observations can be fused in a sequential fashion. For example, the observations \(O_1, O_2, \dots, O_N\) can be fused simultaneously with one large fusion step, but it is also possible to fuse the observations in a sequential fashion. This sequential fusion proceeds as follows. Let \(\Pr_0\) be the prior probability distribution of \(H\). Now fuse the single observation \(O_1\) to get the posterior probability distribution \(\Pr_1\). To fuse on observation \(O_2\), the prior distribution \(\Pr_0\) for \(H\) should be replaced with \(\Pr_1\), and then the single observation \(O_2\) should be fused using the new prior. This process continues until all of \(O_1, O_2, \dots, O_N\) have been fused. Fusing observations in a sequential manner also provides a means of performing context specific fusion with a computational complexity of \(O(N)\). The computational complexity's dependence on \(M\), the domain size of the hypothesis variable, depends on the subtype of credal set used. In the case of point probabilities however, the computational complexity with respect to \(M\) is \(O(M)\). The overall computational complexity for fusing point probability distributions in a context specific manner is \(O(NM)\).

\subsection{General Fusion using point probabilities}\label{sec:probability_g}

This section describes the general fusion process when the credal sets consist of single probability distributions. For now, assume that each \(S_i\) is a single probability distribution with respective probabilities \(p_{i,1}, p_{i,2}, \dots, p_{i,M}\), where \(p_{i,j}\) is the prior probability that \(H_i = j\). With probability distributions, it is required that \(\sum_{j=1}^M p_{i,j} = 1\). Since it is required that \(H_1 = H_2 = \dots = H_N (= H)\), we know that \(E = 1\). 
%This only leaves as being possible the states: 
%\[(H_1, H_2, \dots, H_N, E) = (1, 1, \dots, 1, 1); (2, 2, \dots, 2, 1); \dots ; (M, M, \dots, M, 1)\] 
%The respective prior probabilities of these states are respectively:
%\[\forall j \in \{1, 2, \dots, M\} : \Pr(j, j, \dots, j, 1) = \prod_{i=1}^N p_{i,j}\]
%After renormalization, the posterior probabilities are respectively:
%\[\forall j' \in \{1, 2, \dots, M\} : \Pr(j', j', \dots, j', 1) = \frac{\prod_{i=1}^N p_{i,j'}}{\sum_{j=1}^M\prod_{i=1}^N p_{i,j}}\]
%The fused model \(S_\bullet\) is now the probability distribution defined by the above posterior probabilities.
Bayes' rule gives gives the following posterior probability distribution for \(H\):
\[\forall j' \in \{1, 2, \dots, M\} : \Pr(H_1 = j'|H_1 = H_2 = \dots = H_N) = \frac{\prod_{i=1}^N p_{i,j'}}{\sum_{j=1}^M\prod_{i=1}^N p_{i,j}}\]

\begin{framed} %************************** begin example ***********************
As an example of general fusion, consider the same machine from the context specific fusion example: a machine that can be in one of 2 states: ``functional"; or ``non-functional". The machine's state is the hypothesis variable \(H\), and the \(M = 2\) states are respectively enumerated by 1  and 2. Again there are \(N = 3\) sensors in place to determine the state of the machine, but instead of these sensors simply returning a direct observation, they instead return their own guess at the probability distribution for \(H\). 

Assume that sensor 1 returns a 45\% probability of \(H_1 =1\) (the machine is functional); sensor 2 returns a 60\% probability of \(H_2 = 1\); and sensor 3 returns a 10\% probability of \(H_3 = 1\). Since it is known that \(H_1 = H_2 = H_3\) (which is equivalent to requiring that \(E = 1\)), the posterior probability distribution for \(H\), the common value, is:
\[\Pr(H = 1) = \frac{\Pr(H_1 = H_2 = H_3 = 1)}{\Pr(H_1 = H_2 = H_3)} = \frac{(0.45)(0.6)(0.1)}{(0.45)(0.6)(0.1) + (0.55)(0.4)(0.9)} = 0.12\]
\[\Pr(H = 2) = \frac{\Pr(H_1 = H_2 = H_3 = 2)}{\Pr(H_1 = H_2 = H_3)} = \frac{(0.55)(0.4)(0.9)}{(0.45)(0.6)(0.1) + (0.55)(0.4)(0.9)} = 0.88\]

There is additional complexity to the sensors since they now have to return probability distributions as opposed to raw data. Unlike context specific fusion however, prior probability values and conditional probabilities do not have to be accumulated ahead of time (except possibly for the purpose of ``calibrating" each sensor to return probabilities).  
\end{framed} % *************************** end example **************************

Similar to context specific fusion, credal sets can also be fused in a sequential fusion. Let credal sets \(S_1, S_2, \dots, S_N\) cover the hypothesis variable \(H\). \(S_1, S_2, \dots, S_N\) can be fused in a single large fusion step, but it is also possible to fuse these credal sets in a sequential fashion as follows: \(S_1\) and \(S_2\) are fused to form \(S'_2\); then \(S'_2\) and \(S_3\) are fused to form \(S'_3\); and so on. Again, sequential fusion allows general fusion to proceed with a computational complexity of \(O(N)\) with respect to \(N\). The computational complexity with respect to \(M\) depends on the subtype of credal set used, but for point probabilities the computational complexity is again \(O(M)\) with respect to \(M\). The overall computational complexity for fusing point probability distributions in a general manner is \(O(NM)\).

The following sections will now focus on fusion where the credal set subtype is a set of probability distributions, as opposed to a single probability distribution.

% ****************************************** Fusion using complex uncertainty structures ***********************

\section{Fusion using nontrivial credal sets}\label{sec:Complex}

This section will describe both context specific and general fusion using credal sets that denote sets of probability distributions as opposed to single probability distributions. These credal sets are referred to as being ``nontrivial". 

The basic idea of generalizing fusion to nontrivial credal sets, is that the output credal set should contain every possible probability distribution that results from fusion using probability distributions chosen from each input credal set. Ideally, the output credal set should denote as small a set as possible. Details related to context specific and general fusion are given in the next sections.

\subsection{Context Specific Fusion using Nontrivial Credal Sets}

A high level description of the process of context specific fusion using credal sets can be found in \cite{[Zaffalon2002],[Karlsson2011]}.

For context specific fusion, a ``prior" credal set of the chosen subtype \(S_0\) is needed that describes \(H\). For each observation \(O_i\) (\(i = 1, 2, \dots, N\)), let \(o_i\) denote the value assigned to \(O_i\). For each \(j = 1, 2, \dots, M\), let \(P_{i,j}\) denote the set of all possible values of \(\Pr(O_i = o_i|H = j)\). The resultant credal set \(S_\bullet\) should now satisfy the following \textbf{containment property}: 

\begin{defn}
\textbf{The Containment Property for Context Specific Fusion:} 

First, choose an arbitrary probability distribution \(p_1, p_2, \dots, p_M\) from \(S_0\). For each \(i = 1, 2, \dots, N\), and \(j = 1, 2, \dots, M\) consider an arbitrary choice of probability \(p_{i,j}\) where \(p_{i,j} \in P_{i,j}\). The probability distribution given by:
\[\forall j' = 1, 2, \dots, M : p_{\bullet,j'} = \frac{p_{j'}\prod_{i=1}^N p_{i,j'}}{\sum_{j=1}^M p_j\prod_{i=1}^N p_{i,j}}\] 
should now be contained by the posterior credal set \(S_\bullet\), no matter the choice of \(p_j\)'s and \(p_{i,j}\)'s. \(S_\bullet\) is considered ``tight" if no other probability distributions are contained. \(S_\bullet\) is considered ``maximally tight" if there is no other credal set of the \emph{same subtype} \(S_\bullet'\) that satisfies the containment property and is a proper subset of \(S_\bullet\).
\end{defn}

Like with point probabilities, context specific fusion using nontrivial credal sets can proceed in a sequential manner. However, the resultant credal set may not be as tight as the credal set that results from simultaneous fusion.

It is also important to note that \textbf{the cost of acquiring each \(P_{i,j}\) will not be counted as part of our analysis of the computational complexity} of various algorithms for context specific fusion.

\subsection{General Fusion using Nontrivial Credal Sets}

General fusion using credal sets is described in \cite{[Karlsson2011]}. 

For general fusion, the resultant credal set \(S_\bullet\) should satisfy the following \textbf{containment property}:

\begin{defn}\label{def:containment_property_general_fusion}
\textbf{The Containment Property for General Fusion:}

Choose arbitrary probability distributions \(\Pr_1, \Pr_2, \dots, \Pr_N\) from \(S_1, S_2, \dots, S_N\) respectively. The resultant probability distribution from fusing \(\Pr_1, \Pr_2, \dots, \Pr_N\) should be contained by \(S_\bullet\), no matter the choice of \(\Pr_i\)'s. \(S_\bullet\) is considered ``tight" if no other probability distributions are contained. \(S_\bullet\) is considered ``maximally tight" if there is no other credal set of the \emph{same subtype} \(S_\bullet'\) that satisfies the containment property and is a proper subset of \(S_\bullet\).
\end{defn}

Like with point probabilities, general fusion using nontrivial credal sets can proceed in a sequential manner. However, the resultant credal set may not be as tight as the structure that results from simultaneous fusion.
 
When the causal network from Figure \ref{fig:Fusion_Causal_Network}(b) is treated as a credal network, the strong extension \cite{[Cozman2000]} bears a similarity to the containment property. It is important to note however, that the ``strong extension" of credal networks requires tightness, something that is not a requirement of the containment property. Also in the context of this paper, the maximally tight posterior credal set of the correct subtype may not be as tight as the convex hull that constitutes the strong extension.

\subsection{Lower and Upper Probability Bounds}

As noted in \cite{[DeCampos1994], [Zaffalon1999], [Zaffalon2002]}, determining the lower and upper bounds for posterior probabilities requires the simultaneous minimization and maximization of probabilities. Let \(A\) denote a condition of interest, and let \(B\) denote a condition that is forced to be true (\(B\) denotes \(\forall i = 1, 2, \dots, N : O_i = o_i\) in the case of context specific fusion, and \(B\) denotes \(E = 1\) in the case of general fusion). If \(\PrL\) and \(\PrU\) denote lower and upper probability bounds respectively, then (\cite{[DeCampos1994]}): 
\[\PrL(A|B) = \frac{\PrL(A \wedge B)}{\PrL(A \wedge B) + \PrU(\neg A \wedge B)} \quad\text{and}\quad \PrU(A|B) = \frac{\PrU(A \wedge B)}{\PrU(A \wedge B) + \PrL(\neg A \wedge B)}\]
For the credal set subtypes considered by this paper, the minimization (maximization) of \(\Pr(A \wedge B)\) does not interfere or interact with the maximization (minimization) of \(\Pr(\neg A \wedge B)\).

\subsection{Approximate approaches}

In many cases, credal sets are denoted by listing their ``extreme points". When credal sets are denoted by listing their extreme points, they are not confined to any subtype such as probability interval distributions or Dempster-Shafer models. The paper \cite{[Karlsson2011]} states and proves a theorem (referred to in \cite{[Karlsson2011]} as Theorem 2) that implies that context specific and general fusion using credal sets can be done exactly, meaning that the containment property is satisfied and that the resultant credal set is ``tight". This is not necessarily the case if the credal sets are restricted to a specific subtype. 

In the context of this paper, the output credal set has the same subtype as the credal sets used for the prior data in the case of context specific fusion, and the same subtype as the input credal sets in the case of general fusion. However, due to the limitations on the expressive power of each subtype of credal set, it is rarely possible to return a credal set of the desired subtype that is ``tight". 
Moreover, in many cases, finding the tightest possible output credal set of the desired subtype may be computationally intractable, as will be seen in the subsequent sections. 
Both of these limitations imply that most fusion approaches discussed here will not return a tight credal set of the desired subtype. However, all fusion approaches will satisfy the containment property, something that Dempster's rule of combination (section \ref{sec:conjunctive_combination}) fails to satisfy.

In sections \ref{sec:prob_intervals_credal_sets} and \ref{sec:DS_credal_sets}, probability interval distributions and Dempster-Shafer models are shown to be a more memory efficient alternative to listing the extreme points of credal sets. This increase in memory efficiency is argued to compensate for the decrease in accuracy caused when non-tight credal sets of the desired subtype are returned by fusion.

% ************************************** Fusion with probability intervals ********************************

\section{Probability Interval Fusion}\label{sec:probability_intervals}

The use of probability intervals as opposed to point probabilities is discussed in \cite{[DeCampos1994], [Guo2010]} and \cite[section 4]{[Walley1996]}. 

\begin{defn}
A \textbf{probability interval distribution} \(S\) over the values \(1, 2, \dots, M\) is a set of closed intervals \([l_1, u_1], [l_2, u_2], \dots, [l_M, u_M]\). A probability distribution \(p_1, p_2, \dots, p_M\) is contained by \(S\) if and only if \(\forall j = 1, 2, \dots, M : l_j \leq p_j \leq u_j\). In addition, the lower and upper bounds of the intervals must satisfy the following properties: 
\begin{align*}
(\text{The intervals must be subsets of \([0,1]\)}) \quad & \forall j = 1, 2, \dots, M: \; 0 \leq l_j \leq u_j \leq 1 \\
(\text{At least one probability distribution is contained}) \quad & \sum_{j=1}^M l_j \leq 1 \leq \sum_{j=1}^M u_j \\
(\text{All bounds are reachable}) \quad & \forall j' = 1, 2, \dots, M: \; l_{j'} \geq 1 - \sum_{j : j \neq j'} u_j \\
& \forall j' = 1, 2, \dots, M: \; u_{j'} \leq 1 - \sum_{j : j \neq j'} l_j
\end{align*}

An important restriction on the bounds of the probability intervals, is that for any bound, the bound can be reached by at least one probability distribution contained by \(S\). Let \(p_1, p_2, \dots, p_M\) be an arbitrary probability distribution contained by \(S\). Consider \(p_{j'}\). Aside from the lower bound of \(l_{j'}\), \(p_{j'}\) is also limited by the bounds placed on the other probabilities since \(p_{j'} = 1 - \sum_{j : j \neq j'} p_j\). Setting all other probabilities to their maximum values creates another lower bound for \(p_{j'}\): \(1 - \sum_{j : j \neq j'} u_j\). For \(p_{j'}\) to attain the value \(l_{j'}\), it must be the case that \(l_{j'} \geq 1 - \sum_{j : j \neq j'} u_j\). A similar argument provides a restriction on the upper bound of \(p_{j'}\).
\end{defn}

\subsection{Probability Intervals and credal sets}\label{sec:prob_intervals_credal_sets}

This section will give a simple example that demonstrates how a probability interval distribution can have a large number of extreme points, which makes the style of representation that is commonly used for credal sets, listing the extreme points, computationally intractable. Although it is known in \cite{[Tessem1992]} that the number of extreme points in a probability interval distribution is large, a concrete simple example is provided here for the convenience of the reader. This subsection will give an example of a probability interval distribution \(S\) over the values \(1, 2, \dots, M\), for which the number of extreme points is \(\Omega(2^M/M^2)\). In other words, the number of extreme points is exponential with respect to \(M\). 

Let \(M\) be even. Let the \(j^\text{th}\) probability interval be \([l_j, u_j] = [0, 2/M]\). An extreme probability distribution of \(S\) is formed by choosing \(M/2\) values from \(1, 2, \dots, M\) to be assigned a probability of \(2/M\), and all other probabilities are assigned \(0\). The number of extreme probability distributions is hence:
\[{M \choose M/2} = \frac{M!}{(M/2)!^2}\] 
using \(\ln(n!) \in [n\ln(n) - n + 1, (n+1)\ln(n) - n + 1]\) gives:
\begin{align*}
\frac{M!}{(M/2)!^2} = & \exp(\ln(M!) - 2\ln((M/2)!)) \\
\geq & \exp((M\ln(M) - M + 1) - 2((M/2+1)\ln(M/2) - M/2 + 1)) \\
= & \exp(M\ln(M) - (M+2)\ln(M/2) - 1) %\\
%= & \exp(M\ln(M) - (M+2)(\ln(M)-\ln(2)) - 1) \\
%= & \exp(M\ln(2) - 2\ln(M) + 2\ln(2) - 1) 
= \frac{2^M \cdot 4}{M^2 \cdot e} \\
\in & \Omega(2^M/M^2)
\end{align*}

Here, big-``Omega" notation is used to denote a lower-bound (the opposite of big-``O" notation). A probability interval distribution requires the storage of \(O(M)\) values, while the credal set requires the storage of \(\Omega(2^M/M^2)\) extreme probability distributions.

With this example, it is clear that representing a probability interval distribution by its extreme points is not efficient from a memory perspective, and is hence also inefficient from a time perspective. For instance if \(M = 20\), then the number of extreme points is \({20 \choose 10} = 184756\).

%While this example demonstrates the memory efficiency of using a probability interval distribution verses an exhaustive listing of the extreme points, it is important to note that not every credal set can be represented using a probability interval distribution.

\subsection{Context Specific Fusion with Probability Intervals}\label{sec:intervals_context_specific_fusion} %********************** Context Specific Fusion, Intervals **********************

A high level description of the process of context specific fusion using probability intervals can be found in \cite[section 4, calculus]{[Walley1996]}, and \cite{[Zaffalon2002]}. 

Let \(S_0\) and \([l_1, u_1], [l_2, u_2], \dots, [l_M, u_M]\) denote the prior probability interval distribution for \(H\).

After the observations \(O_i = o_i\) have been received for each \(i = 1, 2, \dots, N\), for each \(j = 1, 2, \dots, M\), the set of possible values of \(\Pr(O_i = o_i | H = j)\) is an interval \(P_{i,j} = [l_{i,j}, u_{i,j}]\). Note that for each \(i = 1, 2, \dots, N\), that the intervals \([l_{i,1}, u_{i,1}], [l_{i,2}, u_{i,2}], \dots, [l_{i,M}, u_{i,M}]\) \emph{do not} collectively form a probability interval distribution.

The posterior probability interval distribution for \(H\) is determined by computing the smallest and largest possible posterior probabilities for each value of \(H\). Let this posterior distribution be denoted by \([l_{\bullet,1}, u_{\bullet,1}], [l_{\bullet,2}, u_{\bullet,2}], \dots, [l_{\bullet,M}, u_{\bullet,M}]\). 

To find these extremes, let \(p_1, p_2, \dots, p_M\) denote an arbitrary prior probability distribution for \(H\) that is contained by \(S_0\), and let \(p_{i,j}\) for each \(i = 1, 2, \dots, N\) and \(j = 1, 2, \dots, M\) denote an arbitrary probability from the interval \([l_{i,j}, u_{i,j}]\). 

For an arbitrary \(j' = 1, 2, \dots, M\), in order to compute \(l_{\bullet,j'}\), the probability of \(H = j' \wedge \forall i \in \{1, 2, \dots, N\} : O_i = o_i\) should be minimized, while the probability of \(H \neq j' \wedge \forall i \in \{1, 2, \dots, N\} : O_i = o_i\) should be maximized. 
This can be done by setting \(p_{j'} = l_{j'}\); \(p_{i,j'} = l_{i,j'}\) for each \(i = 1, 2, \dots, N\); and \(p_{i,j} = u_{i,j}\) for each \(i = 1, 2, \dots, N\) and \(j = 1, 2, \dots, M\) where \(j \neq j'\). 
To decide upon each \(p_j\) where \(j \neq j'\), a greedy maximization approach is used. Each \(p_j\) is set to \(l_j\) by default, and the following process is repeated: Find \(j \in \{1, 2, \dots, M\} \setminus \{j'\}\) that maximizes \(c_j = \prod_{i=1}^N p_{i,j}\). Next, \(p_j\) should be set to the highest allowed probability (the probability is limited by both \(u_j\) and the fact that \(\sum_{j=1}^M p_j = 1\)). \(j\) should then be removed from the set \(j \in \{1, 2, \dots, M\} \setminus \{j'\}\), and a new \(j\) should be chosen. This process repeats until \(\sum_{j=1}^M p_j = 1\). A similar process is used to compute each \(u_{\bullet,j'}\).
 
The following algorithm depicts the process of context specific fusion using probability intervals. To save space, the steps involved in computing the upper bounds \(u_{\bullet,j'}\) will be shown in parentheses beside the steps for computing the lower bounds \(l_{\bullet,j'}\). 

\vspace{5mm}

\begin{algorithmic}
\FOR{\(j = 1\) to \(M\)}
	\STATE \(l_{\Pi,j} \gets \prod_{i=1}^N l_{i,j}\)
	\STATE \(u_{\Pi,j} \gets \prod_{i=1}^N u_{i,j}\)
\ENDFOR
\FOR{\(j' = 1\) to \(M\)}
	\STATE \texttt{// \(l_{\bullet,j'}\) (\(u_{\bullet,j'}\)) will be computed.}
	\FOR{\(j = 1\) to \(M\)}
		\STATE \(p_j \gets l_j\) (\(p_j \gets u_j\))
		\IF{\(j = j'\)}
			\STATE \texttt{/* The prior probability of \(\Pr(H = j' \wedge \forall i = 1, 2, \dots, N : O_i = o_i)\) should be minimized (maximized). */}
			\STATE \(c_j \gets l_{\Pi,j}\) (\(c_j \gets u_{\Pi,j}\))
			\STATE \(b_j \gets 0\)
		\ELSE
			\STATE \texttt{/* The prior probability of \(\Pr(H \neq j' \wedge \forall i = 1, 2, \dots, N : O_i = o_i)\) should be maximized (minimized). */}
			\STATE \(c_j \gets u_{\Pi,j}\) (\(c_j \gets l_{\Pi,j}\))
			\STATE \(b_j \gets 1\)
		\ENDIF
	\ENDFOR
	\STATE \(\sigma \gets \sum_{j=1}^M l_j\) (\(\sigma \gets \sum_{j=1}^M u_j\))
	\WHILE{\(\sigma < 1\) (\(\sigma > 1\))}
		\STATE Find the \(j\) where \(b_j = 1\) that maximizes \(c_j\).
		\STATE \(p_j \gets p_j + \min(u_j - l_j, 1 - \sigma)\) (\(p_j \gets p_j - \min(u_j - l_j, \sigma - 1)\))
		\STATE \(\sigma \gets \sigma + \min(u_j - l_j, 1 - \sigma)\) (\(\sigma \gets \sigma - \min(u_j - l_j, \sigma - 1)\))
		\STATE \(b_j \gets 0\)
	\ENDWHILE
	\STATE \(l_{\bullet,j'} \gets \frac{p_{j'} \cdot c_{j'}}{\sum_{j=1}^M p_j \cdot c_j}\) (\(u_{\bullet,j'} \gets \frac{p_{j'} \cdot c_{j'}}{\sum_{j=1}^M p_j \cdot c_j}\))
\ENDFOR
\end{algorithmic}

\vspace{5mm}

The overall time complexity for context specific fusion using probability intervals is \(O(NM + M^2)\).

\begin{framed} %****************************************** Begin Example (Intervals, Context Specific)************************
As an example of context specific fusion using probability intervals, the same example used for point probabilities in section \ref{sec:probability_cs} will be used. This time, however a \(\pm 0.05\) margin will be included on each probability:
\begin{align*}
\Pr(H = 1) &= [0.85, 0.95]                                                          & \Pr(H = 2) &= [0.05, 0.15] \\
\Pr(O_1 = \text{``low temperature"}|H = 1) &= [0.85, 0.95]  & \Pr(O_1 = \text{``low temperature"}|H = 2) &= [0.35, 0.45] \\
\Pr(O_2 = \text{``low load"}|H = 1) &= [0.25, 0.35]               & \Pr(O_2 = \text{``low load"}|H = 2) &= [0.55, 0.65] \\
\Pr(O_3 = \text{``low current"}|H = 1) &= [0.65, 0.75]          & \Pr(O_3 = \text{``low current"}|H = 2) &= [0.15, 0.25] 
\end{align*} 

If sensor 1 returns \(O_1 = \)``high temperature"; sensor 2 returns \(O_2 = \)``low load"; and sensor 3 returns \(O_3 = \)``low current"; then the posterior probability interval distribution for \(H\) is:
\[\Pr(H = 1|O_1, O_2, O_3) \approx [0.3036, 0.9428] \quad \Pr(H = 2|O_1, O_2, O_3) \approx [0.0572, 0.6964]\]
By comparison with the example from \ref{sec:probability_cs}, it can be seen that the containment property is holding.
\end{framed} %******************************************** End Example *************************

\subsection{General Fusion with Probability Intervals}\label{sec:intervals_general_fusion} %********************** General Fusion, Intervals **********************

Each credal set \(S_i\) is a probability interval distribution, and the prior probability of \(H_i = j\) is a closed interval \([l_{i,j}, u_{i,j}]\) instead of the point probability \(p_{i,j}\). \(S_i\) now describes a set of probability distributions as opposed to a single probability distribution. 

Here it should be note that exact fusion is not possible as probability intervals do not have the necessary expressive power to denote the exact set of possible fused probability distributions. Two approaches to approximate fusion will be covered in the next two subsections:

Finding the maximally tight posterior probability interval distribution for general fusion requires an algorithm for solving the following NP-hard problem:

\begin{prob}\label{prob:sum_prod}
\textbf{Optimum sum of products}

\textbf{Input:} 

Two positive integers \(n\) and \(m\).

Two \(n \times m\) arrays of non-negative real numbers: \(a_{i,j}\) and \(b_{i,j}\) for each \(i = 1, 2, \dots, n\) and \(j = 1, 2, \dots, m\). It must be the case that: 
\[\forall i \in \{1, 2, \dots, n\} : \forall j \in \{1, 2, \dots, m\} : 0 \leq a_{i,j} \leq b_{i,j}\]

One \(n\) length vector of non-negative real numbers: \(c_i\) for each \(i = 1, 2, \dots, n\). It must be the case that:
\[\forall i \in \{1, 2, \dots, n\} : \sum_{j=1}^m a_{i,j} \leq c_i \leq \sum_{j=1}^m b_{i,j}\]
In addition, a choice between maximization and minimization must be made.

\textbf{Internal Variables to be optimized:}

One \(n \times m\) array of non-negative real numbers: \(x_{i,j}\) for each \(i = 1, 2, \dots, n\) and \(j = 1, 2, \dots, m\). The following restrictions hold:
\begin{align*}
\forall i \in \{1, 2, \dots, n\} : \forall j \in \{1, 2, \dots, m\} : & a_{i,j} \leq x_{i,j} \leq b_{i,j} \\
\forall i \in \{1, 2, \dots, n\} : & \sum_{j=1}^m x_{i,j} = c_i \\ 
\end{align*}  

\textbf{Output:} 

The maximum, or minimum depending on choice, possible value of the expression 
\[\sum_{j=1}^m \prod_{i=1}^n x_{i,j}\]
\end{prob}

Problem \ref{prob:sum_prod} in essence takes \(n\) unnormalized probability interval distributions over a domain of \(m\) values: \([a_{i,1}, b_{i,1}], [a_{i,2}, b_{i,2}], \dots, [a_{i,m}, b_{i,m}]\), and extracts from each an unnormalized probability distribution \(x_{i,1}, x_{i,2}, \dots, x_{i,m}\) that sums to \(c_i\). The probability distributions are chosen to either maximize or minimize the probability of agreement between all chosen probability distributions.   
A proof of the NP-hardness of problem \ref{prob:sum_prod} is given in the Appendix. 

It is not hard to show that the \(x_{i,j}\)'s that optimize \(\sum_{j=1}^m \prod_{i=1}^n x_{i,j}\) attain a ``corner state". That is, for each \(i = 1, 2, \dots, N\), \(x_{i,j} = a_{i,j}\) or \(x_{i,j} = b_{i,j}\) for all but one \(j = 1, 2, \dots, M\). There are a finite number of corner states, so as noted in \cite{[Zaffalon1999],[Zaffalon2002]}, problem \ref{prob:sum_prod} can be solved via an exhaustive search of the corner states. 

Since problem \ref{prob:sum_prod} is NP-hard, approximate solutions are necessary for tractable calculations. None of the approximations made in this paper will violate the containment property. In \cite{[Antonucci2013ECSQARU]}, an optimization problem that encompasses problem \ref{prob:sum_prod} is solved in an approximate manner using hill climbing iterations.

\subsubsection{Approach 1}\label{sec:intervals_general_fusion_1} %********************** General Fusion, Intervals, Approach #1 **********************

If computational intractability is not an issue, problem \ref{prob:sum_prod} can be solved to find the tightest possible lower and upper bounds for the posterior probability distribution. The following algorithm depicts the process of general fusion using probability intervals. To save space, the steps involved in computing the upper bounds \(u_{\bullet,j'}\) will be shown in parentheses beside the steps for computing the lower bounds \(l_{\bullet,j'}\). 

\vspace{5mm}

\begin{algorithmic}
\FOR{\(j' = 1\) to \(M\)}
	\STATE \texttt{// \(l_{\bullet,j'}\) (\(u_{\bullet,j'}\)) will be computed.}
	\STATE \(q \gets \prod_{i=1}^N l_{i,j'}\) (\(q \gets \prod_{i=1}^N u_{i,j'}\))
	\STATE \texttt{// \(q\) is the minimized (maximized) prior probability of \(H_1 = H_2 = \dots = H_N = j'\).}
	\STATE \texttt{/* The maximum (minimum) prior probability of \(H_1 = H_2 = \dots = H_N \neq j'\) will now be computed using problem \ref{prob:sum_prod}: */}
	\STATE \(n \gets N\)
	\STATE \(m \gets M-1\)
	\FOR{\(i = 1\) to \(N\)}
		\STATE \(c_i \gets 1 - l_{i,j'}\) (\(c_i \gets 1 - u_{i,j'}\))
		\FOR{\(j = 1\) to \(M-1\)}
			\IF{\(j < j'\)}
				\STATE \(a_{i,j} \gets l_{i,j}\) and \(b_{i,j} \gets u_{i,j}\)
			\ELSE
				\STATE \(a_{i,j} \gets l_{i,j+1}\) and \(b_{i,j} \gets u_{i,j+1}\)
			\ENDIF
		\ENDFOR
	\ENDFOR
	\STATE Solve problem \ref{prob:sum_prod} with the values of \(n\), \(m\), \(a_{i,j}\), \(b_{i,j}\), \(c_i\), and use maximization (minimization). Assign the result to \(r\).
	\STATE \(l_{\bullet,j'} \gets \frac{q}{q + r}\) (\(u_{\bullet,j'} \gets \frac{q}{q + r}\))
\ENDFOR
\end{algorithmic}

\vspace{5mm}

The overall computational complexity for approach \#1 is \(O(NM + M \cdot f(N, M - 1))\) where \(f(n, m)\) is the computational complexity of problem \ref{prob:sum_prod}. 
%This computational complexity ignores the setup times for each application of problem \ref{prob:sum_prod}. 
While the arrays are being freshly generated for each application of problem \ref{prob:sum_prod} in the pseudo code above, the computational complexity assumes a single array can be pre-calculated at a cost of \(O(NM)\) and used for all applications of problem \ref{prob:sum_prod} with different columns ignored. 
%Moreover, the amount of work generating the input arrays is dominated by \(f(N, M - 1)\).

Problem \ref{prob:sum_prod} is NP-hard, but when \(n = 2\), the problem is greatly simplified. Problem \ref{prob:sum_prod} when \(n = 2\) becomes a bilinear programming problem. The resultant bilinear programming programming problem can be more easily solved, and provides a means through which the probability interval distributions can be fused in a sequential manner. It should be noted however, that the resultant probability interval distribution will not be as tight as the probability interval distribution formed through simultaneous fusion.

When probability interval distributions are fused in a pairwise manner, the computational complexity of approach \#1 reduces to \(O(NM + N \cdot M \cdot f(2, M - 1))\).

\subsubsection{Approach 2}\label{sec:intervals_general_fusion_2} %********************** General Fusion, Intervals, Approach #2 **********************

Theory from \cite{[DeCampos1994]} can be used for the general fusion of probability interval distributions. It should be noted however, that the approach presented here will fail to be maximally tight for the following reasons: In the context of general fusion, the hypothesis variables \(H_1, H_2, \dots, H_N\) are all independent when \(E\) is ignored. When a joint probability interval distribution is formed that covers the hypothesis variables, the independence between \(H_1, H_2, \dots, H_N\) can no longer be enforced. This makes possible joint probability distributions over \(H_1, H_2, \dots, H_N\) that do not satisfy the independence between the hypothesis variables.

When \(E = 1\), \(H\) will denote the common value of \(H_1, H_2, \dots, H_N\).

Ignoring the variable \(E\), the joint probability interval for \(\Pr(\forall i \in \{1, 2, \dots, N\} : H_i = j_i)\) is \(\left[\prod_{i=1}^N l_{i,j_i}, \prod_{i=1}^N u_{i,j_i}\right]\)

For each \(j' \in \{1, 2, \dots, M\}\), the smallest and largest posterior probability \(\Pr(H = j'|E = 1)\) is 
\[l_{\bullet,j'} = \frac{\PrL(H = j' \wedge E = 1)}{\PrL(H = j' \wedge E = 1) + \PrU(H \neq j' \wedge E = 1)}\] and 
\[u_{\bullet,j'} = \frac{\PrU(H = j' \wedge E = 1)}{\PrU(H = j' \wedge E = 1) + \PrL(H \neq j' \wedge E = 1)}\] respectively.

For each \(j' \in \{1, 2, \dots, M\}\), \(\prod_{i=1}^N l_{i,j'}\) is the smallest possible prior probability \(\Pr_L(H = j' \wedge E = 1)\). The maximum prior probability \(\Pr_U(H \neq j' \wedge E = 1)\) seems to be \(\sum_{j : j \neq j'} \prod_{i=1}^N u_{i,j}\). However, while each upper bound is attainable, upper bounds may not be \emph{simultaneously attainable}. Another upper bound on the prior probability \(\Pr(H \neq j' \wedge E = 1)\) arises from a lower bound on the prior probability \(\Pr(H = j' \vee E = 0)\). A lower bound on the prior probability \(\Pr(H = j' \vee E = 0)\) that arises directly from the probability intervals is 
\(\prod_{i=1}^N l_{i,j'} + L\) 
where: 
\[L = \prod_{i=1}^N\sum_{j=1}^M l_{i,j} - \sum_{j=1}^M\prod_{i=1}^N l_{i,j}\]
The maximum prior probability \(\Pr_U(H \neq j' \wedge E = 1)\) is:
\(\min\left(\sum_{j : j \neq j'} \prod_{i=1}^N u_{i,j}, 1 - \prod_{i=1}^N l_{i,j'} - L\right)\)

The smallest possible posterior probability \(\Pr_L(H = j' | E = 1)\) is:
\[l_{\bullet,j'} = \frac{\prod_{i=1}^N l_{i,j'}}{\min\left(\prod_{i=1}^N l_{i,j'} + \sum_{j : j \neq j'} \prod_{i=1}^N u_{i,j}, 1 - L\right)}\]

Using a similar argument, the largest possible posterior probability \(\Pr_U(H = j' | E = 1)\) is:
\[u_{\bullet,j'} = \frac{\prod_{i=1}^N u_{i,j'}}{\max\left(\prod_{i=1}^N u_{i,j'} + \sum_{j : j \neq j'} \prod_{i=1}^N l_{i,j}, 1 - U\right)}\]
where
\[U = \prod_{i=1}^N\sum_{j=1}^M u_{i,j} - \sum_{j=1}^M\prod_{i=1}^N u_{i,j}\]

The above gives the complete approach to computing the posterior probability interval distribution \(S_\bullet\) for \(H\): \([l_{\bullet,1}, u_{\bullet,1}]; [l_{\bullet,2}, u_{\bullet,2}]; \dots; [l_{\bullet,M}, u_{\bullet,M}]\).

The overall computational complexity for approach \#2 is \(O(N \cdot M)\). To achieve this efficiency, the following expressions should be computed in the following order: 
\begin{align*} 
\forall i \in \{1, 2, \dots, N\} :\; & l_{i,\Sigma} = \sum_{j=1}^M l_{i,j} \quad u_{i,\Sigma} = \sum_{j=1}^M u_{i,j} &
\forall j \in \{1, 2, \dots, M\} :\; & l_{\Pi,j} = \prod_{i=1}^N l_{i,j}      \quad u_{\Pi,j} = \prod_{i=1}^N u_{i,j} \\
& L_{\Pi\Sigma} = \prod_{i=1}^N l_{i,\Sigma} \quad U_{\Pi\Sigma} = \prod_{i=1}^N u_{i,\Sigma} &
& L_{\Sigma\Pi} = \sum_{j=1}^M l_{\Pi,j}        \quad U_{\Sigma\Pi} = \sum_{j=1}^M u_{\Pi,j} 
\end{align*}
\begin{align*}
\forall j \in \{1, 2, \dots, M\} :\; & l_{\bullet,j} = \frac{l_{\Pi,j}}{\min(l_{\Pi,j}+(U_{\Sigma\Pi}-u_{\Pi,j}), 1-(L_{\Pi\Sigma}-L_{\Sigma\Pi}))} \\
\forall j \in \{1, 2, \dots, M\} :\; & u_{\bullet,j} = \frac{u_{\Pi,j}}{\max(u_{\Pi,j}+(L_{\Sigma\Pi}-l_{\Pi,j}), 1-(U_{\Pi\Sigma}-U_{\Sigma\Pi}))}
\end{align*}

\begin{framed} %****************************************** Begin Example (Intervals, General Approach #2)************************
As an example of general fusion using approach \#2 for probability intervals, the same example used for point probabilities in section \ref{sec:probability_g} will be used. This time, however a \(\pm 0.05\) margin will be included on each probability:
\begin{align*}
\Pr(H_1 = 1) &= [0.40, 0.50]  & \Pr(H_1 = 2) &= [0.50, 0.60] \\
\Pr(H_2 = 1) &= [0.55, 0.65]  & \Pr(H_2 = 2) &= [0.35, 0.45] \\
\Pr(H_3 = 1) &= [0.05, 0.15]  & \Pr(H_3 = 2) &= [0.85, 0.95] 
\end{align*} 

The posterior probability distribution for \(H\), the common value, is:
\[\Pr(H = 1) \approx [0.0411, 0.2468] \quad \Pr(H = 2) \approx [0.7532, 0.9589]\]
By comparison with the example from \ref{sec:probability_g}, it can be seen that the containment property is holding.
\end{framed} %******************************************** End Example *************************

% ************************************** Fusion with Dempster-Shafer models ********************************
\section{Dempster-Shafer Fusion}\label{sec:DS_Fusion}

A description of Dempster-Shafer theory can be found in \cite[chapter 5]{[Klir2005]} and \cite{[Yager1987]}. 

\begin{defn}
A \textbf{Dempster-Shafer model} \(S\) over the values \(\Val(S) = \{1, 2, \dots, M\}\) is described by a ``mass function" \(m : \Set(S) \rightarrow [0, 1]\) where \(\Set(S) = 2^{\Val(S)} \setminus \{\emptyset\}\). It must be the case that: 
\[\sum_{J \in \Set(S)} m(J) = 1\]
A probability distribution \(p_1, p_2, \dots, p_M\) is contained by \(S\) if and only if 
\[\forall J' \subseteq \{1, 2, \dots, M\}: \sum_{J \subseteq J' \wedge J \neq \emptyset} m(J) \leq \sum_{j \in J'} p_j \leq \sum_{J \cap J' \neq \emptyset \wedge J \neq \emptyset} m(J)\]
In other words, the probability of the outcome \(j\) being a member of \(J'\) is bounded from below by the ``belief":
\[\Bel(J') = \sum_{J \subseteq J' \wedge J \neq \emptyset} m(J)\]
and from above by the ``plausibility": 
\[\Pl(J') = \sum_{J \cap J' \neq \emptyset \wedge J \neq \emptyset} m(J)\]
Any probability distribution contained by \(S\) can be generated in the following manner: For each \(J \in \Set(S)\), the weight contained by \(m(J)\) is partitioned between the elements of \(J\). Every and only the probability distributions contained by \(S\) can be formed from this process.
\end{defn}

Dempster-Shafer models have a greater expressive power than probability intervals. Every probability interval distribution has an equivalent Dempster-Shafer model, but only a small fraction of Dempster-Shafer models have an equivalent probability interval distribution.

In a manner similar to the use of probability intervals, a Dempster-Shafer model can be completely characterized by the lower bound ``belief function" \(\Bel: \Set(S) \rightarrow [0, 1]\). The belief function must satisfy the following properties:

\begin{align*}
& (\text{The belief/lower bound must be contained by \([0, 1]\)}) \\
& \quad \forall J \in \Set(S) : 0 \leq \Bel(J) \leq 1 \\ 
& (\text{The lower bounds must respect the union of disjoint sets}) \\ 
& \quad \forall J_1, J_2 \in \Set(S) : J_1 \cap J_2 = \emptyset \implies \Bel(J_1 \cup J_2) \geq \Bel(J_1) + \Bel(J_2) \\
& (\text{The lower bound must be 1 for the entire domain}) \\ 
& \quad \Bel(\Val(S)) = 1 \\
\end{align*}

A Dempster-Shafer model can also be completely characterized by the upper bound ``plausibility function" \(\Pl: \Set(S) \rightarrow [0, 1]\). The belief function must satisfy the following properties:

\begin{align*}
& (\text{The plausibility/upper bound must be contained by \([0, 1]\)}) \\
& \quad \forall J \in \Set(S) : 0 \leq \Pl(J) \leq 1 \\ 
& (\text{The upper bounds must respect the union of disjoint sets}) \\ 
& \quad \forall J_1, J_2 \in \Set(S) : J_1 \cap J_2 = \emptyset \implies \Pl(J_1 \cup J_2) \leq \Pl(J_1) + \Pl(J_2) \\
& (\text{The upper bound must be 1 for the entire domain}) \\ 
& \quad \Pl(\Val(S)) = 1 \\
\end{align*}

Given a valid belief/lower bound function or a valid plausibilty/upper bound function, the mass function can be computed via the inclusion/exclusion principle \cite[pg. 4]{[Yager2008Book]}:
\[\forall J' \in \Set(S) : m(J') = \sum_{J \subseteq J' \wedge J \neq \emptyset} (-1)^{|J'|+|J|}\Bel(J)\]
\[\forall J' \in \Set(S) : m(J') = \sum_{J \supseteq (\Val(S) \setminus J') \wedge J \neq \emptyset} (-1)^{1+|\Val(S)|+|J'|+|J|}\Pl(J)\]

%From here on, Dempster-Shafer models will be described via their lower bound function instead of by their mass function.

\subsection{Dempster-Shafer models and credal sets}\label{sec:DS_credal_sets}

Like with probability intervals, a Dempster-Shafer model \(S\) over the domain \(1, 2, \dots, M\) for which the number of extreme points greatly exceeds the size of \(S\) will be constructed to prove the utility of Dempster-Shafer models in comparison with listing the extreme points. The size of the Dempster-Shafer model is \(O(2^M)\). In this case, the number of extreme points will be \(\Omega(M!)\). Let \(M \gg 1\) be arbitrary: For each \(J \in \Set(S)\), let \(m(J) = \frac{1}{2^M - 1}\). An extreme probability distribution is formed by choosing a permutation of \(1, 2, \dots, M\). Let \(\rho : \Val(S) \rightarrow \Val(S)\) denote this permutation. All probability mass gravitates to \(\rho(1)\); followed by \(\rho(2)\); and so on. The probability assigned to \(\rho(j)\) for each \(j = 1, 2, \dots, M\) is: \(p_j = \frac{2^{(M-j)}}{(2^M) - 1}\). The number of permutations \(\rho\) is \(M!\), so the number of extreme points is \(\Omega(M!)\). 

Again, it is clear that representing a Dempster-Shafer model by its extreme points is not computationally efficient. For instance, if \(M = 20\), a Dempster-Shafer model requires \(2^{20} - 1 = 1048575\) values, while the number of extreme points is \(20! \approx 2.4329 \times 10^{18}\).

%************************* Context Specific Fusion, Dempster-Shafer ****************************
\subsection{Context Specific Fusion with Dempster-Shafer models}\label{sec:DS_context_specific_fusion}

An approach to context specific fusion using Dempster-Shafer models is given in \cite{[Delmotte2004]}. The approach presented here however, will differ from the approach given in \cite{[Delmotte2004]}, since the presented approach will aim to satisfy the containment property. The approach presented here will bear similarities to the approach from \cite{[Zaffalon2002]}.

Context specific fusion using Dempster-Shafer models proceeds in a very similar manner to context specific fusion using probability intervals. 

Let \(S_0\); \(m_0 : \Set(H) \rightarrow [0, 1]\); and \(\Bel_0 : \Set(H) \rightarrow [0, 1]\) all denote the prior Dempster-Shafer model for \(H\).

After the observations \(O_i = o_i\) have been received for each \(i = 1, 2, \dots, N\), for each \(j = 1, 2, \dots, M\), the set of possible values of \(\Pr(O_i = o_i | H = j)\) is an interval \(P_{i,j} = [l_{i,j}, u_{i,j}]\). Each \(P_{i,j}\) is simply an interval, as a full Dempster-Shafer model that covers \(O_i\) is not required. Only the scenario of \(O_i = o_i\) is under consideration.

The posterior Dempster-Shafer model for \(H\) is determined by computing the smallest (or largest) possible posterior probabilities for each nonempty subset of \(\Val(H)\). Let the lower (or upper) posterior probability of \(J \in \Set(H)\) be denoted by \(\Bel_\bullet(J)\) (or \(\Pl_\bullet(J)\)). 

The following algorithm depicts the process of context specific fusion using probability intervals. To save space, the steps involved in computing the plausibilities \(\Pl_\bullet\) will be shown in parentheses beside the steps for computing the beliefs \(\Bel_\bullet\). (Note however, that only computing the beliefs are necessary for the posterior Dempster-Shafer model.)

\vspace{5mm}

\begin{algorithmic}
\FOR{\(j = 1\) to \(M\)}
	\STATE \(l_{\Pi,j} \gets \prod_{i=1}^N l_{i,j}\)
	\STATE \(u_{\Pi,j} \gets \prod_{i=1}^N u_{i,j}\)
\ENDFOR
\FOR{all \(J' \in \Set(H)\)}
	\STATE \texttt{// \(\Bel_\bullet(J')\) (\(\Pl_\bullet(J')\)) will be computed.}
	\FOR{\(j = 1\) to \(M\)}
		\STATE \(p_j \gets 0\) 
		\IF{\(j \in J'\)}
			\STATE \texttt{/* The prior probability of \(\Pr(H \in J' \wedge \forall i = 1, 2, \dots, N : O_i = o_i)\) should be minimized (maximized). */}
			\STATE \(c_j \gets l_{\Pi,j}\) (\(c_j \gets u_{\Pi,j}\))
		\ELSE
			\STATE \texttt{/* The prior probability of \(\Pr(H \notin J' \wedge \forall i = 1, 2, \dots, N : O_i = o_i)\) should be maximized (minimized). */}
			\STATE \(c_j \gets u_{\Pi,j}\) (\(c_j \gets l_{\Pi,j}\))
		\ENDIF
	\ENDFOR
	\FOR{all \(J \in \Set(H)\)}
		\IF{\(J \subseteq J'\)}
			\STATE Find the \(j \in J\) that minimizes (maximizes) \(c_j\).
		\ELSIF{\(J \cap J' = \emptyset\)}
			\STATE Find the \(j \in J\) that maximizes (minimizes) \(c_j\).
		\ELSE
			\STATE Find the \(j \in J \setminus J'\) (\(j \in J \cap J'\)) that maximizes \(c_j\).
		\ENDIF
		\STATE \(p_j \gets p_j + m_0(J)\)
	\ENDFOR
	\STATE \(\Bel_\bullet(J') \gets \frac{\sum_{j' \in J'} p_{j'} \cdot c_{j'}}{\sum_{j=1}^M p_j \cdot c_j}\) (\(\Pl_\bullet(J') \gets \frac{\sum_{j' \in J'} p_{j'} \cdot c_{j'}}{\sum_{j=1}^M p_j \cdot c_j}\))
\ENDFOR
\end{algorithmic}

\vspace{5mm}

The overall time complexity for context specific fusion (including the cost of determining the final masses using the inclusion/exclusion principle) using Dempster-Shafer models is \(O(NM + 2^{2M})\). Since the size \(D\) of the prior Dempster-Shafer model for \(H\) is approximately \(2^M\), the time complexity is in fact \(O(NM + D^2)\). %Moreover, to save space in a Dempster-Shafer model, sets \(J \in 2^{\Val(H)} \setminus \{\emptyset\}\) for which \(m_0(J) = 0\) can be omitted, leaving a list of \(\langle J, m_0(J) \rangle\) pairs. The amount of \(\langle J, m_0(J) \rangle\) pairs may be vastly less that \(2^M\). Moreover, this memory reduction directly  

\begin{framed} %****************************************** Begin Example (Dempster-Shafer, Context Specific)************************
As an example of context specific fusion using Dempster-Shafer models, the same example used for point probabilities in section \ref{sec:probability_cs} will be used. A \(\pm 0.05\) margin will be included to form the prior Dempster-Shafer model for \(H\) and each probability interval for the observed evidence:
\begin{align*}
m_0(\{1\}) &= 0.85                     & m_0(\{2\}) &= 0.05 &
m_0(\{1, 2\}) &= 0.1 \\
\Pr(o_1|H = 1) &= [0.05, 0.15]  & \Pr(o_1|H = 2) &= [0.55, 0.65] \\
\Pr(o_2|H = 1) &= [0.25, 0.35]  & \Pr(o_2|H = 2) &= [0.55, 0.65] \\
\Pr(o_3|H = 1) &= [0.65, 0.75]  & \Pr(o_3|H = 2) &= [0.15, 0.25] 
\end{align*} 

The posterior Dempster-Shafer model for \(H\) is:
\[m_\bullet(\{1\}) \approx 0.3036 \quad m_\bullet(\{2\}) \approx 0.0572 \quad m_\bullet(\{1,2\}) \approx 0.6392\]
By comparison with the example from \ref{sec:probability_cs}, it can be seen that the containment property is holding.
\end{framed} %******************************************** End Example *************************

%************************* General Fusion, Dempster-Shafer ****************************
\subsection{General Fusion with Dempster-Shafer models}\label{sec:DS_general_fusion}

Dempster's rule of combination (described in \cite{[Yager1987]}) performs general fusion of Dmpster-Shafer models. However, Dempster's rule of combination fails to satisfy the containment property.

Each structure \(S_i\) is a Dempster-Shafer model. \(S_i\) is denoted by either the mass function \(m_i : \Set(H_i) \rightarrow [0, 1]\); or the belief function \(\Bel_i : \Set(H_i) \rightarrow [0, 1]\). Also, the resultant Dempster-Shafer model \(S_\bullet\) is denoted by either \(m_\bullet : \Set(H) \rightarrow [0, 1]\); or \(\Bel_\bullet : \Set(H) \rightarrow [0, 1]\).

Finding the tightest possible Dempster-Shafer model for general fusion requires an algorithm for solving the following problem:

\begin{prob}\label{prob:mass_choice}
\textbf{Optimum sum of products, Dempster-Shafer variant}

\textbf{Input:} 

Two positive integers \(n\) and \(m\).

A set \(A\) with \(m\) distinct quantities.

An \(n\) length array of functions: \(f_i : 2^A \setminus \{\emptyset\} \rightarrow [0, +\infty)\) for each \(i = 1, 2, \dots, n\).

In addition, a choice between maximization and minimization must be made.

\textbf{Internal Variables to be optimized:}

An \(n\) length array of functions: \(g_i : 2^A \setminus \{\emptyset\} \rightarrow A\) for each \(i = 1, 2, \dots, n\). The following restriction must hold:
\[\forall i \in \{1, 2, \dots, n\} : \forall J \in 2^A \setminus \{\emptyset\} : g_i(J) \in J\]

An \(n \times m\) array of non-negative real numbers: \(x_{i,j}\) for each \(i = 1, 2, \dots, n\) and \(j \in A\) where:
\[\forall i \in \{1, 2, \dots, n\} : \forall j \in A : x_{i,j} = \sum_{J \in 2^A \setminus \{\emptyset\} \wedge g_i(J) = j} f_i(J)\]

\textbf{Output:} 

The maximum, or minimum depending on choice, possible value of the expression 
\[\sum_{j \in A} \prod_{i=1}^n x_{i,j}\]
\end{prob}

Problem \ref{prob:mass_choice}, which is directly analogous to problem \ref{prob:sum_prod}, in essence takes \(n\) unnormalized Dempster-Shafer models over a domain of \(m\) values: \(A = \{1, 2, \dots, m\}\). From each Dempster-Shafer model \(i = 1, 2, \dots, n\), each probability mass is focused onto a single element, forming an unnormalized probability distribution \(x_{i,1}, x_{i,2}, \dots, x_{i,m}\). The probability distributions are chosen to either maximize or minimize the probability of agreement between all chosen probability distributions.

\subsubsection{Approach 1}\label{sec:DS_general_fusion_1} %********************** General Fusion, Dempster-Shafer **********************

Like with probability intervals, if computational intractability is not an issue, problem \ref{prob:mass_choice} can be solved to find the tightest Dempster-Shafer model for the posterior probability distribution. The following algorithm depicts the process of general fusion using Dempster-Shafer models. To save space, the steps involved in computing the plausibilities \(\Pl_\bullet\) will be shown in parentheses beside the steps for computing the beliefs \(\Bel_\bullet\). 
(Note however, that only computing the beliefs are necessary for the posterior Dempster-Shafer model.)

\vspace{5mm}

\begin{algorithmic}
\FOR{all \(J' \in \Set(H)\)}
	\STATE \texttt{// \(\Bel_\bullet(J')\) (\(\Pl_\bullet(J')\)) will be computed.}
	\STATE \texttt{/* The minimum prior probability of \(E = 1 \wedge H \in J'\) (\(E = 1 \wedge H \notin J'\)) will now be computed using problem \ref{prob:mass_choice}: */}
	\STATE \(n \gets N\)
	\STATE \(m \gets |J'|\) (\(m \gets M - |J'|\))
	\STATE \(A \gets J'\) (\(A \gets \Val(H) \setminus J'\))
	\FOR{\(i = 1\) to \(n\)}
		\FOR{all \(J \in 2^A \setminus \{\emptyset\}\)}
			\STATE \(f_i(J) \gets m_i(J)\)
		\ENDFOR
	\ENDFOR
	\STATE Solve problem \ref{prob:mass_choice} with the values of \(n\), \(m\), \(A\), \(f_i\), and use minimization. Assign the result to \(q\).
	\STATE \texttt{/* The maximum prior probability of \(E = 1 \wedge H \notin J'\) (\(E = 1 \wedge H \in J'\)) will now be computed using problem \ref{prob:mass_choice}: */}
	\STATE \(m \gets M - |J'|\) (\(m \gets |J'|\))
	\STATE \(A \gets \Val(H) \setminus J'\) (\(A \gets J'\))
	\FOR{\(i = 1\) to \(n\)}
		\FOR{all \(J \in 2^A \setminus \{\emptyset\}\)}
			\STATE \texttt{/* Since the prior probability of \(E = 1 \wedge H \in A\) is being maximized, probability mass gravitates into \(A\): */}
			\STATE \(f_i(J) \gets \sum_{J'' \in \Set(H) \wedge J'' \cap A = J} m_i(J'')\)
		\ENDFOR
	\ENDFOR
	\STATE Solve problem \ref{prob:mass_choice} with the values of \(n\), \(m\), \(A\), \(f_i\), and use maximization. Assign the result to \(r\).
	\STATE \(\Bel_\bullet(J') \gets \frac{q}{q+r}\) (\(\Pl_\bullet(J') \gets \frac{r}{q+r}\))	
\ENDFOR
\end{algorithmic}

\vspace{5mm}

The overall computational complexity for approach \#1 is \(O(N \cdot 2^{2M} + 2^M \cdot g(N, M))\) where \(g(n, m)\) is the computational complexity of problem \ref{prob:mass_choice}. 
%This computational complexity ignores the setup times for each application of problem \ref{prob:mass_choice}. 
While the input functions are being freshly generated for each application of \ref{prob:mass_choice} in the pseudo code above, the computational complexity assumes that all input functions can be pre-calculated at a cost of \(O(N \cdot 2^{2M})\) and used for all applications of problem \ref{prob:mass_choice} with different entries ignored. 
%Moreover, the amount of work generating the input functions is dominated by \(g(N, M)\).

When Dempster-Shafer models are fused in a pairwise manner, the computational complexity of approach \#1 reduces to \(O(N \cdot 2^{2M} + N \cdot 2^M \cdot g(2, M))\).

\subsubsection{Approach 2}\label{sec:DS_general_fusion_2} %********************** General Fusion, Dempster-Shafer **********************

The second approach to general fusion using Dempster-Shafer models also uses theory from \cite{[DeCampos1994]} and is similar to general fusion approach \#2 for probability intervals. Like with probability intervals, a joint Dempster-Shafer model is created that covers the variables \(H_1, H_2, \dots, H_N\). Again, like with probability intervals, the joint Dempster-Shafer model will fail to enforce the independence between variables \(H_1, H_2, \dots, H_N\). For this fusion approach the assumption that \(N \geq 2\) is important.    

The joint Dempster-Shafer model for the variables \(H_1, H_2, \dots, H_N\), denoted by \(S_\times\), is created as follows: consider \(J_\times = J_1 \times J_2 \times \dots \times J_N\) for arbitrary \(J_1, J_2, \dots, J_N \in \Set(H)\). Let the mass assigned to \(J_\times\) be: \(m_\times(J_\times) = m_1(J_1) m_2(J_2) \dots m_n(J_N)\). For any \(J_\times \in \Set(\{H_1, H_2, \dots, H_N\})\), if there does not exist any \(J_1, J_2, \dots, J_N \in \Set(H)\) such that \(J_\times = J_1 \times J_2 \times \dots \times J_N\), then the mass assigned to \(J_\times\) is 0: \(m_\times(J_\times) = 0\). 

The calculation of \(\Bel_\bullet(J')\) (and \(\Pl_\bullet(J')\)) for an arbitrary \(J' \in \Set(H)\) will now be the focus. When point probabilities are used, the posterior probability for \(H \in J'\) is: \(\Pr(H \in J') = \frac{\Pr(H \in J' \wedge E = 1)}{\Pr(H \in J' \wedge E = 1) + \Pr(H \notin J' \wedge E = 1)}\). The condition that \(E = 1\) requires that \(H_1 = H_2 = \dots = H_N\), and \(H\) denotes the common value. As noted in section \ref{sec:Complex}, 
\[\PrL(H \in J'|E = 1) = \frac{\PrL(H \in J' \wedge E = 1)}{\PrL(H \in J' \wedge E = 1) + \PrU(H \notin J' \wedge E = 1)}\] 
and
\[\PrU(H \in J'|E = 1) = \frac{\PrU(H \in J' \wedge E = 1)}{\PrU(H \in J' \wedge E = 1) + \PrL(H \notin J' \wedge E = 1)}\]

Therefore:
\begin{align*}
\forall J' \in \Set(H): \quad & \Bel_\bullet(J') = \frac{\Bel_\times(H \in J' \wedge E = 1)}{\Bel_\times(H \in J' \wedge E = 1) + \Pl_\times(H \in (\Val(H) \setminus J') \wedge E = 1)} \\
\forall J' \in \Set(H): \quad & \Pl_\bullet(J') = \frac{\Pl_\times(H \in J' \wedge E = 1)}{\Pl_\times(H \in J' \wedge E = 1) + \Bel_\times(H \in (\Val(H) \setminus J') \wedge E = 1)} \\
\end{align*}
where 
\begin{align*}
\forall J' \in \Set(H): \quad & \Bel_\times(H \in J' \wedge E = 1) = \sum_{j \in J'}\prod_{i=1}^N m_i(\{j\}) \\
\forall J' \in \Set(H): \quad & q_\times(H \in J' \wedge E = 1) = \prod_{i=1}^N \sum_{J \supseteq J'} m_i(J) \\
\forall J' \in \Set(H): \quad & \Pl_\times(H \in J' \wedge E = 1) = \sum_{J \subseteq J' \wedge J \neq \emptyset} (-1)^{1+|J|} q_\times(H \in J \wedge E = 1) \\
%\sum_{\rr{J_1, J_2, \dots, J_N \in \Set(H)}{J' \cap J_1 \cap J_2 \cap \dots \cap J_N \neq \emptyset}}\prod_{i=1}^N m_i(J_i) \\
\end{align*}

Note that the quantities \(\Bel_\times(H \in \emptyset \wedge E = 1)\) and \(\Pl_\times(H \in \emptyset \wedge E = 1)\) default to 0.

The expression \(\sum_{j \in J'}\prod_{i=1}^N m_i(\{j\})\) is non-zero if and only if there exists some \(j \in J'\) for which \(m_i(\{J\}) > 0\) for all \(i = 1, 2, \dots, N\). For this approach to general fusion using Dempster-Shafer models, masses assigned to singleton elements of \(\Set(H)\) are important for the creation of non-trivial Dempster-Shafer models. 

The computational complexity of approach \#2 is \(O(N \cdot 2^{2M})\). 
%The bulk of the computational complexity is found in the evaluation of \(\Pl_\times(H \in J' \wedge E = 1)\).   

\begin{framed} %****************************************** Begin Example (Dempster-Shafer, General Approach #2)************************
As an example of general fusion using approach \#2 for Dempster-Shafer models, the same example used for point probabilities in section \ref{sec:probability_g} will be used. This time however, a \(\pm 0.05\) margin will be included to form each Dempster-Shafer model:
\begin{align*}
m_1(\{1\}) &= 0.40 & m_1(\{2\}) &= 0.50 & m_1(\{1,2\}) = 0.10 \\
m_2(\{1\}) &= 0.55 & m_2(\{2\}) &= 0.35 & m_2(\{1,2\}) = 0.10 \\
m_3(\{1\}) &= 0.05 & m_3(\{2\}) &= 0.85 & m_3(\{1,2\}) = 0.10 \\ 
\end{align*} 

The posterior probability distribution for \(H\), the common value, is:
\[m_\bullet(\{1\}) \approx 0.0411 \quad m_\bullet(\{2\}) \approx 0.7532 \quad m_\bullet(\{1,2\}) \approx 0.2057\]
By comparison with the example from \ref{sec:probability_g}, it can be seen that the containment property is holding.
\end{framed} %******************************************** End Example *************************

\subsection{Dempster's Rule of Combination}\label{sec:conjunctive_combination}

Dempster's rule of combination performs general fusion of Dempster Shafer models. Dempster's rule of combination, described in \cite{[Yager1987]}, proceeds as follows:
\[\forall J' \in \Set(H) : m_\bullet(J') = \frac{1}{K}\sum_{\begin{array}{c}
J_1, J_2, \dots, J_N \in \Set(H) \\
J_1 \cap J_2 \cap \dots \cap J_N = J' \\
\end{array}} m_1(J_1)m_2(J_2) \dots m_N(J_N)\]
where \(K\) is a normalization constant that ensures that \(\sum_{J' \in \Set(H)} m_\bullet(J') = 1\).

As will be shown in the following example, Dempster's rule of combination fails to satisfy the containment property. The example is from \cite{[Eastwood-Yanush-chapter-2015]}.

\begin{framed} %***************** Dempster's rule of combination fails to satisfy the containment property. ****************
As an example of Dempster's rule of combination failing to satisfy the containment property, let \(N = 2\) and \(M = 2\). 
Let Dempster-Shafer model \(S_1\) be defined by:
\[m_1(\{1\}) = 0.1 \quad m_1(\{2\}) = 0.1 \quad m_1(\{1, 2\}) = 0.8\] 
Let Dempster-Shafer model \(S_2\) be the same: \(S_2 = S_1\).

Dempster's rule of combination gives the following resultant Dempster-Shafer model \(S_\bullet\):
\[m_\bullet(\{1\}) \approx 0.1735 \quad m_\bullet(\{2\}) \approx 0.1735 \quad m_\bullet(\{1, 2\}) \approx 0.6531\]

Now consider probability distribution \(\Pr_1 \in S_1\):
\[{\Pr}_1(1) = 0.1 \quad {\Pr}_1(2) = 0.9\]
Let probability distribution \(\Pr_2 \in S_2\) be the same: \(\Pr_2 = \Pr_1\).

Fusing \(\Pr_1\) and \(\Pr_2\) gives \(\Pr_\bullet\):
\[{\Pr}_\bullet(1) \approx 0.0122 \quad {\Pr}_\bullet(2) \approx 0.9878\]

It is readily apparent that \(\Pr_\bullet \notin S_\bullet\), which violates the containment property for general fusion. 
This example demonstrates that Dempster's rule of combination violates the containment property for general fusion.  
\end{framed}

Due to the fact that Dempster's rule of combination fails to satisfy the containment property, general fusion approach \#2 is proposed as an alternative to Dempster's rule of combination.

%************************************************ Conclusion ******************************************
\section{Conclusion}\label{sec:Conclusion}

This paper has given a taxonomy of approaches to both context specific and general fusion using both probability interval distributions and Dempster-Shafer models. Fusion approaches that were covered include:
\begin{itemize}
\item Point probability distributions: 
\begin{itemize}
\item Context specific fusion (section \ref{sec:probability_cs}): The computational complexity is \(O(NM)\).
\item General fusion (section \ref{sec:probability_g}): The computational complexity is \(O(NM)\).
\end{itemize}
\item Probability Interval Distributions:
\begin{itemize}
\item Context specific fusion (section \ref{sec:intervals_context_specific_fusion}): The computational complexity is \(O(NM + M^2)\), and the posterior is maximally tight.
\item General fusion approach \#1 (section \ref{sec:intervals_general_fusion_1}): The computational complexity is \(O(NM + Mf(N,M-1))\), and the posterior is maximally tight (\(f(n,m)\) is the complexity of problem \ref{prob:sum_prod}).
\item General fusion approach \#2 (section \ref{sec:intervals_general_fusion_2}): The computational complexity is \(O(NM)\), and the posterior is not maximally tight.
\end{itemize}
\item Dempster-Shafer models:
\begin{itemize}
\item Context specific fusion (section \ref{sec:DS_context_specific_fusion}): The computational complexity is \(O(NM + 2^{2M})\), and the posterior is maximally tight.
\item General fusion approach \#1 (section \ref{sec:DS_general_fusion_1}): The computational complexity is \(O(N \cdot 2^{2M} + 2^Mg(N,M))\), and the posterior is maximally tight (\(g(n,m)\) is the complexity of problem \ref{prob:mass_choice}).
\item General fusion approach \#2 (section \ref{sec:DS_general_fusion_2}): The computational complexity is \(O(N \cdot 2^{2M})\), and the posterior is not maximally tight.
\end{itemize}
\end{itemize}

The containment property, which requires that the fusion of any choice of point probability distributions be contained in the resultant credal set, is presented as an objective requirement that all fusion approaches should satisfy. Dempster's rule of combination is shown to not satisfy the containment property (see section \ref{sec:conjunctive_combination}).

Credal sets are convex sets of probability distributions, and a typical approach to denoting credal sets is to list their extreme points. It has been shown in \cite{[Karlsson2011]} that context specific fusion and general fusion can be exactly and computationally efficiently performed by listing the extreme points of credal sets. Exact fusion requires that the containment property holds and that the resultant model is tight. This at first seems to imply that listing the extreme points of credal sets are the optimal approach to describing convex sets of probability distributions. This paper shows however, that representing probability interval distributions and Dempster-Shafer models using lists of their extreme points can lead to excessive memory requirements and poor computational efficiency. Therefore, this paper proposes probability intervals and Dempster-Shafer models as a computationally tractable alternative to the listing of extreme points. 

Unlike listing extreme points, context specific and general fusion using probability interval distributions and Dempster-Shafer models can rarely be performed exactly. Moreover, probability intervals and Dempster-Shafer models lack the expressive power to denote a tight posterior credal set. All approaches to fusion proposed here satisfy the containment property, and the approaches presented have varying levels of speed and accuracy. 

There are many directions for future work. Problems \ref{prob:sum_prod} and \ref{prob:mass_choice} can be further investigated for more accurate and computationally efficient algorithms despite problem \ref{prob:sum_prod} being NP-hard. The algorithms for context specific fusion and general fusion can be generalized to ``credal networks" (existing work on credal networks can be found in \cite{[Antonucci2013ECSQARU], [Antonucci2013], [Cano2007], [Cozman2005]}). In addition, the presented approaches can be investigated for specific applications of sensor fusion.

\section*{Acknowledgments}

\begin{small}
This project was partially supported by Natural Sciences and Engineering Research Council of Canada; and the Government of Alberta (Queen Elizabeth II  Scholarship). The authors would also like to thank Drs. D. Sezer and Y. Zinchenko for their valuable assistance.
\end{small}

%\section*{References}

\bibliographystyle{elsarticle-harv}
\bibliography{References}

\begin{thebibliography}{30}
\expandafter\ifx\csname natexlab\endcsname\relax\def\natexlab#1{#1}\fi
\providecommand{\url}[1]{\texttt{#1}}
\providecommand{\href}[2]{#2}
\providecommand{\path}[1]{#1}
\providecommand{\DOIprefix}{doi:}
\providecommand{\ArXivprefix}{arXiv:}
\providecommand{\URLprefix}{URL: }
\providecommand{\Pubmedprefix}{pmid:}
\providecommand{\doi}[1]{\href{http://dx.doi.org/#1}{\path{#1}}}
\providecommand{\Pubmed}[1]{\href{pmid:#1}{\path{#1}}}
\providecommand{\bibinfo}[2]{#2}
\ifx\xfnm\relax \def\xfnm[#1]{\unskip,\space#1}\fi
%Type = Inproceedings
\bibitem[{Antonucci et~al.(2013a)Antonucci, De~Campos, Huber and
  Zaffalon}]{[Antonucci2013ECSQARU]}
\bibinfo{author}{Antonucci, A.}, \bibinfo{author}{De~Campos, C.P.},
  \bibinfo{author}{Huber, D.}, \bibinfo{author}{Zaffalon, M.},
  \bibinfo{year}{2013}a.
\newblock \bibinfo{title}{Approximating credal network inferences by linear
  programming}, in: \bibinfo{booktitle}{European Conference on Symbolic and
  Quantitative Approaches to Reasoning and Uncertainty},
  \bibinfo{publisher}{Springer}, \bibinfo{address}{Berlin Heidelberg}. pp.
  \bibinfo{pages}{13--24}.
%Type = Inproceedings
\bibitem[{Antonucci et~al.(2013b)Antonucci, Huber, Zaffalon, Luginbuhl, Chapman
  and Ladouceur}]{[Antonucci2013]}
\bibinfo{author}{Antonucci, A.}, \bibinfo{author}{Huber, D.},
  \bibinfo{author}{Zaffalon, M.}, \bibinfo{author}{Luginbuhl, P.},
  \bibinfo{author}{Chapman, I.}, \bibinfo{author}{Ladouceur, R.},
  \bibinfo{year}{2013}b.
\newblock \bibinfo{title}{{CREDO}: a military decision-support system based on
  credal networks}, in: \bibinfo{booktitle}{Proceedings of the 16th
  International Conference on Information Fusion}, pp.
  \bibinfo{pages}{1942--1949}.
%Type = Article
\bibitem[{Cano et~al.(2007)Cano, Gomez, Moral and Abellan}]{[Cano2007]}
\bibinfo{author}{Cano, A.}, \bibinfo{author}{Gomez, M.},
  \bibinfo{author}{Moral, S.}, \bibinfo{author}{Abellan, J.},
  \bibinfo{year}{2007}.
\newblock \bibinfo{title}{Hill-climbing and branch-and-bound algorithms for
  exact and approximate inference in credal networks}.
\newblock \bibinfo{journal}{International Journal of Approximate Reasoning}
  \bibinfo{volume}{44}, \bibinfo{pages}{261--280}.
%Type = Article
\bibitem[{Cozman(2000)}]{[Cozman2000]}
\bibinfo{author}{Cozman, F.G.}, \bibinfo{year}{2000}.
\newblock \bibinfo{title}{Credal networks}.
\newblock \bibinfo{journal}{Artificial intelligence} \bibinfo{volume}{120},
  \bibinfo{pages}{199--233}.
%Type = Article
\bibitem[{Cozman(2005)}]{[Cozman2005]}
\bibinfo{author}{Cozman, F.G.}, \bibinfo{year}{2005}.
\newblock \bibinfo{title}{Graphical models for imprecise probabilities}.
\newblock \bibinfo{journal}{International Journal of Approximate Reasoning}
  \bibinfo{volume}{39}, \bibinfo{pages}{167--184}.
%Type = Article
\bibitem[{De~Campos et~al.(1994)De~Campos, Huete and Moral}]{[DeCampos1994]}
\bibinfo{author}{De~Campos, L.M.}, \bibinfo{author}{Huete, J.F.},
  \bibinfo{author}{Moral, S.}, \bibinfo{year}{1994}.
\newblock \bibinfo{title}{Probability intervals: a tool for uncertain
  reasoning}.
\newblock \bibinfo{journal}{International Journal of Uncertainty, Fuzziness and
  Knowledge-Based Systems} \bibinfo{volume}{2}, \bibinfo{pages}{167--196}.
%Type = Article
\bibitem[{Delmotte and Smets(2004)}]{[Delmotte2004]}
\bibinfo{author}{Delmotte, F.}, \bibinfo{author}{Smets, P.},
  \bibinfo{year}{2004}.
\newblock \bibinfo{title}{Target identification based on the transferable
  belief model interpretation of {Dempster-Shafer} model}.
\newblock \bibinfo{journal}{IEEE Trans. Systems, Man and Cybernetics, Part A:
  Systems and Humans} \bibinfo{volume}{34}, \bibinfo{pages}{457--471}.
%Type = Unpublished
\bibitem[{Dezert and Smarandache(2003)}]{[Dezert2003]}
\bibinfo{author}{Dezert, J.}, \bibinfo{author}{Smarandache, F.},
  \bibinfo{year}{2003}.
\newblock \bibinfo{title}{On the generation of hyper-powersets for the {DSmT}}.
\newblock \bibinfo{note}{ArXiv preprint math/0309431}.
%Type = Inproceedings
\bibitem[{Dezert et~al.(2006)Dezert, Tchamova, Smarandache and
  Konstantinova}]{[Dezert2006]}
\bibinfo{author}{Dezert, J.}, \bibinfo{author}{Tchamova, A.},
  \bibinfo{author}{Smarandache, F.}, \bibinfo{author}{Konstantinova, P.},
  \bibinfo{year}{2006}.
\newblock \bibinfo{title}{Target type tracking with {PCR5} and {Dempster's}
  rules: a comparative analysis}, in: \bibinfo{booktitle}{9th International
  Conference on Information Fusion}, \bibinfo{organization}{IEEE}.
%Type = Incollection
\bibitem[{Eastwood and Yanushkevich(2016)}]{[Eastwood-Yanush-chapter-2015]}
\bibinfo{author}{Eastwood, S.C.}, \bibinfo{author}{Yanushkevich, S.N.},
  \bibinfo{year}{2016}.
\newblock \bibinfo{title}{Risk assessment in authentication machines}, in:
  \bibinfo{editor}{Abielmona, R.}, \bibinfo{editor}{Falcon, R.},
  \bibinfo{editor}{Zincir-Heywood, N.}, \bibinfo{editor}{Abbas, H.} (Eds.),
  \bibinfo{booktitle}{Recent Advances in Computational Intelligence in Defense
  and Security, Springer series on Studies in Computational Intelligence}.
  \bibinfo{publisher}{Springer}, pp. \bibinfo{pages}{391--420}.
%Type = Article
\bibitem[{Guo and Tanaka(2010)}]{[Guo2010]}
\bibinfo{author}{Guo, P.}, \bibinfo{author}{Tanaka, H.}, \bibinfo{year}{2010}.
\newblock \bibinfo{title}{Decision making with interval probabilities}.
\newblock \bibinfo{journal}{European Journal of Operational Research}
  \bibinfo{volume}{203}, \bibinfo{pages}{444--454}.
%Type = Phdthesis
\bibitem[{Karlsson(2010)}]{[Karlsson2010]}
\bibinfo{author}{Karlsson, A.}, \bibinfo{year}{2010}.
\newblock \bibinfo{title}{Evaluating credal set theory as a belief framework in
  high-level information fusion for automated decision-making}.
\newblock Ph.D. thesis. Orebro University.
%Type = Article
\bibitem[{Karlsson et~al.(2011)Karlsson, Johansson and Andler}]{[Karlsson2011]}
\bibinfo{author}{Karlsson, A.}, \bibinfo{author}{Johansson, R.},
  \bibinfo{author}{Andler, S.F.}, \bibinfo{year}{2011}.
\newblock \bibinfo{title}{Characterization and empirical evaluation of
  {Bayesian} and credal combination operators}.
\newblock \bibinfo{journal}{Journal of Advances in Information Fusion}
  \bibinfo{volume}{6}, \bibinfo{pages}{150--166}.
%Type = Inproceedings
\bibitem[{Karlsson and Steinhauer(2013)}]{[Karlsson2013]}
\bibinfo{author}{Karlsson, A.}, \bibinfo{author}{Steinhauer, H.J.},
  \bibinfo{year}{2013}.
\newblock \bibinfo{title}{Evaluation of evidential combination operators}, in:
  \bibinfo{booktitle}{Proceedings of the 8th International Symposium on
  Imprecise Probability: Theory and Applications}.
%Type = Article
\bibitem[{Khaleghi et~al.(2013)Khaleghi, Khamis, Karray and
  Razavi}]{[Khaleghi2013]}
\bibinfo{author}{Khaleghi, B.}, \bibinfo{author}{Khamis, A.},
  \bibinfo{author}{Karray, F.O.}, \bibinfo{author}{Razavi, S.N.},
  \bibinfo{year}{2013}.
\newblock \bibinfo{title}{Multisensor data fusion: A review of the
  state-of-the-art}.
\newblock \bibinfo{journal}{Information Fusion} \bibinfo{volume}{14},
  \bibinfo{pages}{28--44}.
%Type = Book
\bibitem[{Klir(2005)}]{[Klir2005]}
\bibinfo{author}{Klir, G.J.}, \bibinfo{year}{2005}.
\newblock \bibinfo{title}{Uncertainty and information: foundations of
  generalized information theory}.
\newblock \bibinfo{publisher}{John Wiley \& Sons}.
%Type = Book
\bibitem[{Sipser(2006)}]{[Sipser2006]}
\bibinfo{author}{Sipser, M.}, \bibinfo{year}{2006}.
\newblock \bibinfo{title}{Introduction to the Theory of Computation, second
  edition}.
\newblock \bibinfo{publisher}{Course Technology, Cengage Learning}.
%Type = Inproceedings
\bibitem[{Smarandache and Dezert(2005a)}]{[Smarandache2005IF]}
\bibinfo{author}{Smarandache, F.}, \bibinfo{author}{Dezert, J.},
  \bibinfo{year}{2005}a.
\newblock \bibinfo{title}{Information fusion based on new proportional conflict
  redistribution rules}, in: \bibinfo{booktitle}{7th International Conference
  on Information Fusion}, \bibinfo{organization}{IEEE}.
%Type = Article
\bibitem[{Smarandache and Dezert(2005b)}]{[Smarandache2005]}
\bibinfo{author}{Smarandache, F.}, \bibinfo{author}{Dezert, J.},
  \bibinfo{year}{2005}b.
\newblock \bibinfo{title}{A simple proportional conflict redistribution rule}.
\newblock \bibinfo{journal}{International Journal of Applied Mathematics and
  Statistics} \bibinfo{volume}{3}.
\newblock \URLprefix \url{http://arxiv.org/abs/cs.AI/0408010}.
%Type = Inproceedings
\bibitem[{Tchamova et~al.(2003)Tchamova, Semerdjiev and
  Dezert}]{[Tchamova2003]}
\bibinfo{author}{Tchamova, A.}, \bibinfo{author}{Semerdjiev, T.},
  \bibinfo{author}{Dezert, J.}, \bibinfo{year}{2003}.
\newblock \bibinfo{title}{Estimation of target behavior tendencies using
  {Dezert-Smarandache} theory}, in: \bibinfo{booktitle}{Proceedings of the 6th
  International Conference on Information Fusion (Fusion 2003)}, pp.
  \bibinfo{pages}{1349--1356}.
%Type = Article
\bibitem[{Tessem(1992)}]{[Tessem1992]}
\bibinfo{author}{Tessem, B.}, \bibinfo{year}{1992}.
\newblock \bibinfo{title}{Interval probability propagation}.
\newblock \bibinfo{journal}{International Journal of Approximate Reasoning}
  \bibinfo{volume}{7}, \bibinfo{pages}{95--120}.
%Type = Inproceedings
\bibitem[{Van~Norden et~al.(2008)Van~Norden, Bolderheij and
  Jonker}]{[Norden2008]}
\bibinfo{author}{Van~Norden, W.}, \bibinfo{author}{Bolderheij, F.},
  \bibinfo{author}{Jonker, C.}, \bibinfo{year}{2008}.
\newblock \bibinfo{title}{Combining system and user belief on classification
  using the {DSmT} combination rule}, in: \bibinfo{booktitle}{11th
  International Conference on Information Fusion, 2008},
  \bibinfo{publisher}{IEEE}.
%Type = Article
\bibitem[{Walley(1996)}]{[Walley1996]}
\bibinfo{author}{Walley, P.}, \bibinfo{year}{1996}.
\newblock \bibinfo{title}{Measures of uncertainty in expert systems}.
\newblock \bibinfo{journal}{Artificial intelligence} \bibinfo{volume}{83},
  \bibinfo{pages}{1--58}.
%Type = Article
\bibitem[{Yager(1987)}]{[Yager1987]}
\bibinfo{author}{Yager, R.R.}, \bibinfo{year}{1987}.
\newblock \bibinfo{title}{On the {Dempster-Shafer} framework and new
  combination rules}.
\newblock \bibinfo{journal}{Information sciences} \bibinfo{volume}{41},
  \bibinfo{pages}{93--137}.
%Type = Article
\bibitem[{Yager(2004)}]{[Yager2004II]}
\bibinfo{author}{Yager, R.R.}, \bibinfo{year}{2004}.
\newblock \bibinfo{title}{On the determination of strength of belief for
  decision support under uncertainty -- {Part II}: fusing strengths of belief}.
\newblock \bibinfo{journal}{Fuzzy Sets and Systems} \bibinfo{volume}{142},
  \bibinfo{pages}{129--142}.
%Type = Article
\bibitem[{Yager and Filev(1995)}]{[Yager1995IEEE]}
\bibinfo{author}{Yager, R.R.}, \bibinfo{author}{Filev, D.P.},
  \bibinfo{year}{1995}.
\newblock \bibinfo{title}{Including probabilistic uncertainty in fuzzy logic
  controller modeling using {Dempster-Shafer} theory}.
\newblock \bibinfo{journal}{IEEE Trans. Systems, Man, and Cybernetics}
  \bibinfo{volume}{25}, \bibinfo{pages}{1221--1230}.
%Type = Book
\bibitem[{Yager and Liu(2008)}]{[Yager2008Book]}
\bibinfo{editor}{Yager, R.R.}, \bibinfo{editor}{Liu, L.} (Eds.),
  \bibinfo{year}{2008}.
\newblock \bibinfo{title}{Classic works of the {Dempster-Shafer} theory of
  belief functions}. volume \bibinfo{volume}{219}.
\newblock \bibinfo{publisher}{Springer}.
%Type = Article
\bibitem[{Yager and Petry(2016)}]{[Yager2016IF]}
\bibinfo{author}{Yager, R.R.}, \bibinfo{author}{Petry, F.},
  \bibinfo{year}{2016}.
\newblock \bibinfo{title}{An intelligent quality-based approach to fusing
  multi-source probabilistic information}.
\newblock \bibinfo{journal}{Information Fusion} \bibinfo{volume}{31},
  \bibinfo{pages}{127--136}.
%Type = Inproceedings
\bibitem[{Zaffalon(1999)}]{[Zaffalon1999]}
\bibinfo{author}{Zaffalon, M.}, \bibinfo{year}{1999}.
\newblock \bibinfo{title}{A credal approach to naive classification}, in:
  \bibinfo{booktitle}{1st International Symposium on Imprecise Probabilities
  and Their Applications}, pp. \bibinfo{pages}{405--414}.
%Type = Article
\bibitem[{Zaffalon(2002)}]{[Zaffalon2002]}
\bibinfo{author}{Zaffalon, M.}, \bibinfo{year}{2002}.
\newblock \bibinfo{title}{The naive credal classifier}.
\newblock \bibinfo{journal}{Journal of statistical planning and inference}
  \bibinfo{volume}{105}, \bibinfo{pages}{5--21}.

\end{thebibliography}

% ********************************************* Appendix **************************************
\appendix
\section{The NP-hardness of Problem \ref{prob:sum_prod}}

The satisfiability problem (SAT) from propositional logic is known to be NP-complete by the Cook-Levin theorem \cite[pg. 276]{[Sipser2006]}.

The formulation of the SAT problem given in \cite[pg. 271]{[Sipser2006]} is: 
\begin{prob}
\textbf{Satifiability (SAT) formulation 1}

\textbf{Input} A propositional formula \(\phi\) of length \(m\), with at most \(n\) binary propositional variables. 

\textbf{Output} A binary yes/no that indicates if there exists an assignment to the \(n\) propositional variables such that \(\phi\) evaluates to ``true".
\end{prob}

Here however, an alternate formulation of the SAT problem is used that is equivalent to the first formulation:
\begin{prob}
\textbf{Satifiability (SAT) formulation 2}

\textbf{Input} A set of \(n\) binary propositional variables \(x_1, x_2, \dots, x_n \in \{0,1\}\).

A set of \(m\) clauses \(\phi_1, \phi_2, \dots, \phi_m\). Each clause is a disjunction: \(\phi_j = l_{j,1} \vee l_{j,2} \vee \dots \vee l_{j,n}\) where \(l_{j,i}\) is either \(F (\text{false})\); \(x_i\); or \(\neg x_i\).

\textbf{Output} A binary yes/no that indicates if there exists an assignment to the \(n\) propositional variables such that every \(\phi_j\) evaluates to ``true".
\end{prob}

Both formulations of SAT are polynomial time reducible to each other.
Formulation 2 can be envisioned as a specific instance of formulation 1 with \(\phi = \phi_1 \wedge \phi_2 \wedge \dots \wedge \phi_m\), and so formulation 2 is readily polynomial time reducible to formulation 1. Formulation 1 can be reduced to formulation 2 in polynomial time via the following process: given the expression tree for \(\phi\), an extra propositional variable can be created for each interior node. A node's dependence on its children can be encoded via a small set of disjunctive clauses. Hence, the condition that \(\phi\) return true can be encoded by a set of disjunctive clauses that can be generated in polynomial time. This set of disjunctive clauses constitutes the polynomial time reduction of formulation 1 to formulation 2.

To establish that problem \ref{prob:sum_prod} is NP-hard, it is sufficient to show that SAT (formulation 2) is polynomial time reducible to problem \ref{prob:sum_prod}. Polynomial time reducible means that SAT can be solved in polynomial time provided that a polynomial time algorithm exists for problem \ref{prob:sum_prod}. SAT can be solved by problem \ref{prob:sum_prod} in the following manner: 

Start with the input to SAT: A set of \(n\) binary propositional variables \(x_1, x_2, \dots, x_n \in \{0,1\}\), and a set of \(m\) clauses \(\phi_1, \phi_2, \dots, \phi_m\). 

SAT is solved via problem \ref{prob:sum_prod} by the following algorithm:

\vspace{5mm}

\begin{algorithmic}
\STATE \(n' \gets n + m\)
\STATE \(m' \gets 2n\)
\FOR{\(i' = 1\) to \(n\)}
	\FOR{\(i = 1\) to \(n\)}
		\IF{\(i = i'\)}
			\STATE \(a_{i',2i-1} \gets 0\) and \(a_{i',2i} \gets 0\)
		\ELSE
			\STATE \(a_{i',2i-1} \gets 1\) and \(a_{i',2i} \gets 1\)
		\ENDIF
		\STATE \(b_{i',2i-1} \gets 1\) and \(b_{i',2i} \gets 1\)
	\ENDFOR
	\STATE \(c_{i'} \gets 2n - 1\)
\ENDFOR
\FOR{\(j = 1\) to \(m\)}
	\FOR{\(i = 1\) to \(n\)}
		\IF{\(l_{j, i} \equiv x_i\)}
			\STATE \(a_{n+j, 2i-1} \gets 0\) and \(a_{n+j, 2i} \gets 1\)
		\ELSIF{\(l_{j, i} \equiv \neg x_i\)}
			\STATE \(a_{n+j, 2i-1} \gets 1\) and \(a_{n+j, 2i} \gets 0\)
		\ELSE
			\STATE \(a_{n+j, 2i-1} \gets 1\) and \(a_{n+j, 2i} \gets 1\)
		\ENDIF
		\STATE \(b_{n+j,2i-1} \gets 1\) and \(b_{n+j,2i} \gets 1\)
	\ENDFOR
	\STATE \(c_{n+j} \gets 2n - 1\)
\ENDFOR 
\STATE Solve problem \ref{prob:sum_prod} with the values of \(n = n'\), \(m = m'\), \(a_{i,j}\), \(b_{i,j}\), \(c_i\), and use maximization. Assign the result to \(r\).
\IF{\(r \geq n\)}
	\STATE \textbf{return:} yes (\(\phi_1 \wedge \phi_2 \wedge \dots \wedge \phi_m\) is satisfiable)
\ELSE
	\STATE \textbf{return:} no (\(\phi_1 \wedge \phi_2 \wedge \dots \wedge \phi_m\) is unsatisfiable)
\ENDIF
\end{algorithmic}

\vspace{5mm}

\begin{figure}[h!]
\begin{centering}
\includegraphics[width = \textwidth]{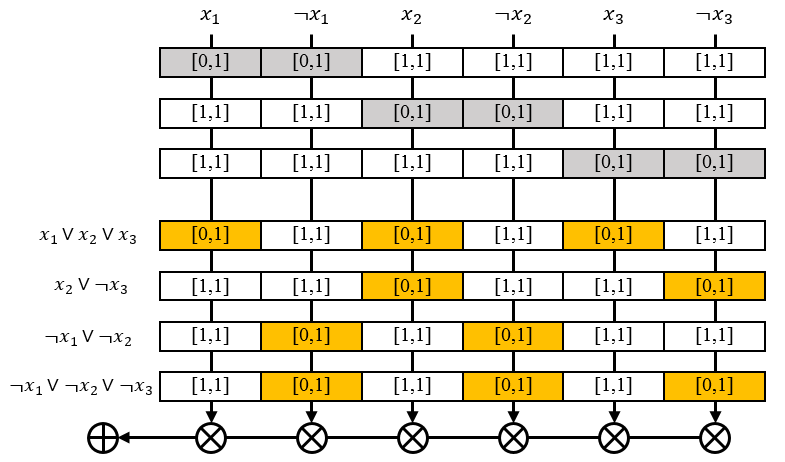}
\caption{A visual depiction of setting up problem \ref{prob:sum_prod} to solve the SAT problem.}
\label{fig:NP_hardness_1}
\end{centering}
\end{figure}

Figure \ref{fig:NP_hardness_1} depicts the use of problem \ref{prob:sum_prod} to solve the SAT problem involving the clauses: \(\phi_1 =         x_1 \vee          x_2 \vee         x_3\); 
\(\phi_2 = F             \vee          x_2 \vee \neg x_3\); 
\(\phi_3 = \neg x_1 \vee \neg x_2 \vee F\); and 
\(\phi_4 = \neg x_1 \vee \neg x_2 \vee \neg x_3\).
Note that problem \ref{prob:sum_prod} is optimized by a ``corner state", wherein the parameters do not take on intermediate values. 
For each of the top \(n\) rows, one of \(x_i\) or \(\neg x_i\) is chosen to be true by forcing the corresponding row entry to 0. The product of the corresponding column is forced to 0. The products of at least \(n\) columns are 0, so the sum of products is at most \(n\). For each of the bottom \(m\) rows, a supporting literal for clause \(\phi_j\) is chosen by again forcing the corresponding row entry to 0. If the chosen supporting literal does not match the choice of \(x_i\)'s in the top \(n\) rows, then another column has a 0 product and the sum of products falls below \(n\). If the clauses are all simultaneously satisfiable, then there exists a choice of assignments to each \(x_i\) and a choice of supporting literal for each \(\phi_j\) so that the product of \(n\) columns is 1, and the sum of products attains a maximum of \(n\).

\end{document}